\documentclass{midl} 

\usepackage{amsmath}
\usepackage{amssymb}
\usepackage{booktabs}
\usepackage{makecell}
\usepackage{bm}
\usepackage{array}
\usepackage{tabu}
\usepackage{multirow}
\usepackage{tabularray}

\usepackage{algorithm}
\usepackage{algpseudocode}
\algrenewcommand\algorithmicthen{}

\usepackage{pgf}

\UseTblrLibrary{booktabs}
\UseTblrLibrary{siunitx}
\NewColumnType{J}[1][]{Q[co=1,1.0,si={table-format=#1,table-number-alignment=center},c]}
\usepackage{siunitx}

\usepackage{graphicx} 

\usepackage{pifont}
\newcommand{\cmark}{\ding{51}}
\newcommand{\xmark}{\ding{55}}

\SetKwInOut{KwIn}{Input}
\SetKwInOut{KwOut}{Output}

\SetKw{Continue}{continue}

\usepackage{makecell}

\usepackage{mwe} 
\jmlrvolume{}
\jmlryear{}
\jmlrpages{}
\jmlrworkshop{Full Paper -- MIDL 2026}

\title[Vesselpose]{Vesselpose: Vessel Graph Reconstruction from Learned Voxel-wise Direction Vectors in 3D Vascular Images}

\midlauthor{
\Name{Rajalakshmi Palaniappan\nametag{$^{1,2}$}}\Email{Rajalakshmi.Palaniappan@mdc-berlin.de}\\
\Name{Christoph Karg\nametag{$^{1}$}}\\
\Name{Nemesio Navarro-Arambula}\nametag{$^{3}$}\\
\Name{Peter Hirsch\nametag{$^{1,4}$}}\orcid{0000-0002-2353-5310}\\
\Name{Kristin Kräker}\nametag{$^{1,5,6,7}$}\\
\Name{Lisa Mais\midlotherjointauthor\nametag{$^{1,3,4}$}}\orcid{0000-0002-9281-2668} \\
\Name{Dagmar Kainmueller\midljointauthortext{Contributed equally}\nametag{$^{1,3,4}$}}\orcid{0000-0002-9830-2415
} \Email{Dagmar.Kainmueller@mdc-berlin.de}\\
\addr $^{1}$ Max-Delbrueck-Center for Molecular Medicine in the Helmholtz Association (MDC), \\
\addr $^{2}$ Humboldt University of Berlin, 
\addr $^{3}$ University of Potsdam, 
\addr $^{4}$ Helmholtz Imaging,\\
\addr $^{5}$ Charité-Universitätsmedizin,
\addr $^{6}$ German Centre for Cardiovascular Research (DZHK), \\
\addr $^{7}$ Experimental and Clinical Research Center (ECRC), a cooperation of Charité-Universitätsmedizin and MDC
}

\begin{document}
\maketitle
 \vspace{-3em}
\begin{abstract}
Blood vessel segmentation and -tracing are essential tasks in many medical imaging applications. Although numerous methods exist, the prevailing segment-then-fix paradigm is fundamentally limited regarding its suitability for modeling the task of complete and topologically accurate vascular network reconstruction. 
Here, we propose an approach to extract  topologically more accurate vascular graphs from 3D image data, building upon highly successful ideas from the related biomedical tasks of cell segmentation and -tracking. Our approach first predicts voxel-wise vessel direction vectors joint with standard vessel segmentation masks. Second, to extract the vascular graph from these predictions, we introduce a direction-vector-guided extension of the TEASAR algorithm.
Our approach achieves state-of-the-art performance on three benchmark datasets, spanning both synthetic and real imagery.
We further demonstrate the applicability of our approach to challenging 3D micro-CT scans of rat heart vasculature.
Finally, we propose meaningful and interpretable measures of topological error, namely false splits and false merges for graphs. Overall, our approach substantially improves the topological accuracy of reconstructed vascular graphs, being able to separate closely apposed vessel segments and handle multiple vascular trees within a single volume.
\end{abstract}

\begin{keywords}
Blood Vessel Reconstruction, Centerline Topology, Evaluation.
\end{keywords}

\section{Introduction}
\label{sec:intro}
Tree-like structures are ubiquitous in living organisms and serve vital functions, for example as blood vessels, airways, or neuronal networks.
Understanding how these branched systems develop, function, and change under pathological conditions requires detailed structural information.
Image-based analysis has become a key tool for investigating such processes at the subcellular to organ scale, using diverse modalities such as MRI, micro-CT, electron microscopy, or light-sheet microscopy~\cite{cheng2024,Walek2023,soubeyrand2023,todorov2020,Obenaus2017}.
To analyze these structures, image data are processed to extract simplified network representations—typically skeletons or graphs—from which features such as overall topology, segment lengths, and branching patterns can be quantified.
For example, \citet{liu2021assessment} showed that altered topology of the microvasculature plays an important role in hepatocellular carcinoma (HCC).
However, obtaining accurate and complete reconstructions remains challenging; manual or semi-automated tracing is still often the method of choice, particularly in cardiovascular imaging \cite{Pampols-Perez2025piezo2,RiosCoronado2025}.

\begin{figure}
    \centering
    \includegraphics[width=0.9\linewidth]{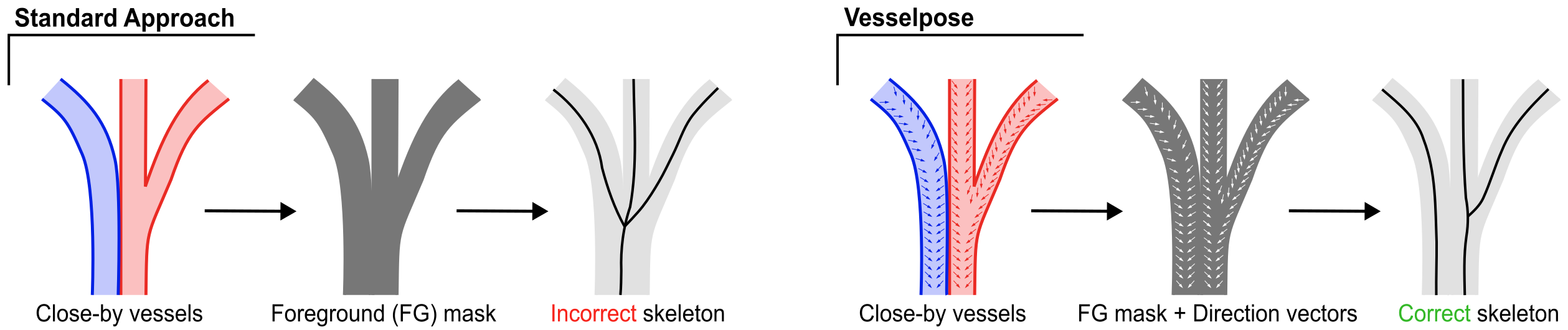}
    \caption{\textbf{Segmentation-and-skeletonization vs.\ Vesselpose.}
Traditional segment-and-skeletonize pipelines often produce incorrect skeletons, especially when distinct vessels lie in close proximity. In contrast, Vesselpose leverages voxel-wise direction vectors to robustly reconstruct vascular trees, naturally handling closely apposed branches as well as multiple distinct trees.}
    \label{fig:our_approach_summary}
    \vspace{-20pt}
\end{figure}

Topological correctness refers to accurately preserving the connectivity of the biological network.
Many vessel-graph extraction pipelines perform foreground-background segmentation of the image first \cite{tetteh2019_deepvesselnet,todorov2020,Wittmann_2025_CVPR}, and then skeletonize the foreground mask using TEASAR or variants of the Lee algorithm \cite{voreen2009,Drees2021voreenSkel,Bumgarner2022vesselvio}.
However, these approaches struggle to achieve topological accuracy: 
Variable imaging contrast can render vessel segments faint or discontinuous, prompting segmentation methods to produce false splits; at the same time, branches running in close proximity often lead to false merges that incorrectly connect distinct vessels.
In the subsequent skeletonization step, such false merge errors create artificial cycles or spurious branching points, as illustrated in the left part of \figureref{fig:our_approach_summary}. 
Topological losses~\cite{lux2025topograph,kirchhoff2024,Shit2021cldice} or simple heuristics like thinning of ground-truth masks may help with this issue to some extent -- However, segmentation as a modular step remains fundamentally ill-suited for modelling the task of topologically correct vessel graph reconstruction.
This limitation is not specific to foreground-background segmentation, but also holds for instance segmentation: 
Although the vasculature forms a globally connected system, imaging typically covers only a restricted anatomical region, yielding multiple disjoint trees; 
while instance segmentation can, in principle, model the separation of different vessel trees, it is not designed to prevent false merges \emph{within} a single tree.

Therefore, alternative strategies are needed to achieve reliable and topologically accurate vascular graph reconstructions.
Earlier work used combinatorial optimization to assemble vessel trees from small centerline tracklets \cite{tueretken2010,tueretken2011}, yielding globally optimal tree reconstructions w.r.t.\ some objective under topological constraints, thereby ensuring topological correctness.
However, even with heuristics and relaxed constraints, Integer Linear Programming (ILP)-based methods \cite{tueretken2016,robben2014,robben2016,rempfler2016minimum} remain computationally expensive and do not scale to large vascular networks.
More recently, image-to-graph frameworks \cite{vesselformer2024,trexplorer2024, trexplorer_super2025}, inspired by DETR \cite{carion2020eccv,zhu2021deformable}, have emerged as a promising direction.
Most of them have been validated primarily on synthetic datasets, where
Vesselformer \cite{vesselformer2024} still produces notable topological errors, while Trexplorer \cite{trexplorer2024} suffers from duplicate branching and premature tracking termination. Trexplorer-Super \cite{trexplorer_super2025} addresses these issues and extends evaluation to real datasets, yet its training and evaluation remain restricted to single-tree structures, whereas real vascular volumes typically contain multiple disjoint trees.
Thus, a more general and computationally feasible solution is still required—one that can robustly extract topologically meaningful graphs from multi-tree vascular networks.

At the same time, issues of topological correctness have been addressed very successfully for the highly related tasks of cell segmentation and tracking in 3D(+t) microscopy data \cite{cellpose2021,Malin-Mayor2023}.
Here, the community has moved away from the traditional segment-then-fix paradigm, replacing deep learning (DL)-based binary segmentation with models that predict pixel-wise shape properties that encode topologically relevant information \cite{hirsch2020_an_auxil_task_for_learn_nucle_segme_in_3d,patchperpix,Sheridan2023lsd}.
Most prominently, Cellpose \cite{cellpose2021,cellpose2022} predicts vector fields pointing toward object centers;
similarly, for cell tracking through time, \citet{Malin-Mayor2023} predict pixel-wise direction vectors that point backward in time to the center of the same or mother cell in the previous frame.
Iteratively following these vectors reconstructs complete cell lineages and ultimately traces each cell back to its origin.
This approach leverages the biological prior that cells divide but do not merge, ensuring a unique predecessor.
These advances highlight the value of predicting pixel-wise topological information, suggesting a promising direction that has not yet been extended to vascular tree reconstruction.

Building on these insights, we propose a method that extracts topologically plausible vessel trees from 3D images using a heuristic solver guided by voxel-wise predictions.
We train a network to predict direction vectors that point toward the vessel centerline while being biased in the rootward direction, leveraging the anatomical prior that vessel diameter typically increases toward the root.
This prior enables robust orientation along the tree and naturally suits vascular and airway networks.
By defining the flow from endpoints toward the root, we circumvent directional ambiguities at branching points, resulting in a well-defined direction vector at each location.

The predicted binary mask and direction vectors then serve as input to a novel skeletonization objective that reconstructs the tree structure by following the learned vector field.
In summary, our main contributions are as follows:
\begin{itemize}
    \item We present a DL-based method that predicts voxel-wise direction vectors from 3D vascular images, which a fast heuristic solver then assembles into a consistent vessel centerline graph.
    \item We introduce meaningful and easily interpretable topology-aware evaluation metrics such as false splits and false merges for graphs, proposing a tailor-made \emph{assignment strategy} (cf.\ \citet{metrics_reloaded2024}) based on hierarchical graph-matching.
    \item We outperform current state-of-the-art on synthetic and real datasets and extend evaluation to a widely used multi-tree dataset \cite{tetteh2019_deepvesselnet} and a real 3D micro-CT dataset, achieving superior topological accuracy and reconstruction quality.
\end{itemize}
The code for the model and evaluation, along with trained models and prediction results, is publicly available at \url{https://github.com/Kainmueller-Lab/Vesselpose}.


\begin{figure}
    \centering
    \includegraphics[width=0.9\linewidth]{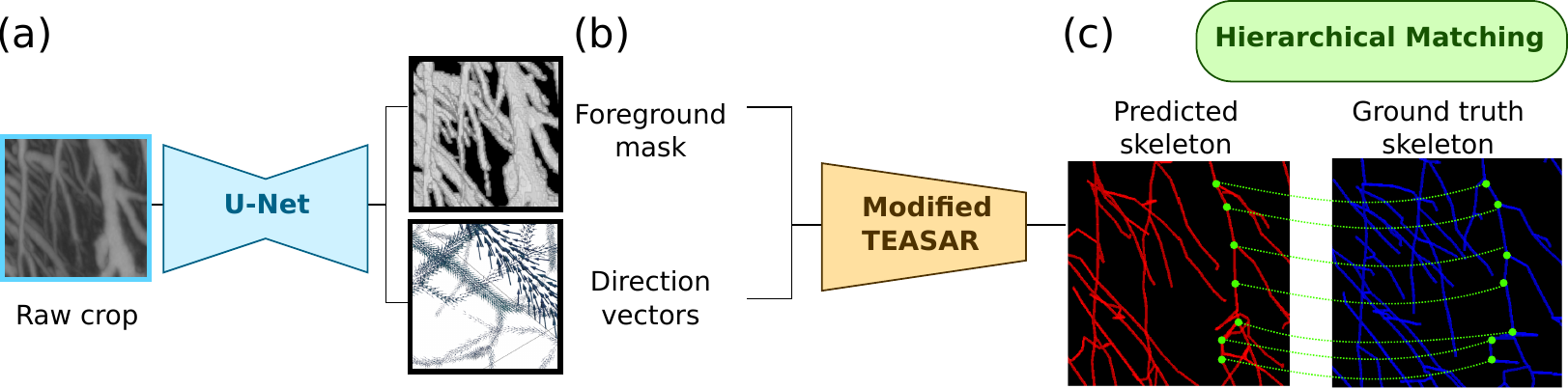}
    \caption{\textbf{Blood vessel reconstruction and evaluation.}
    (a) A U-Net predicts vessel foreground and voxel-wise direction vectors from the raw image.
    (b) A modified TEASAR algorithm extracts a skeleton graph.
    (c)~Predicted skeletons are evaluated against ground-truth using hierarchical graph matching as assignment strategy, which yields topologically meaningful error metrics.
    }
    \label{fig:vessel_tree_prediction}
    \vspace{-15pt}
\end{figure}
\section{Method}
\label{sec:method}
To derive graph representations from raw 3D vascular images, we first predict voxel-wise direction vectors capturing local structure, subsequently assembled into a tree-structured skeleton using a modified TEASAR algorithm~\cite{sato2000teasar}.  \figureref{fig:vessel_tree_prediction} illustrates our approach. 
\sectionref{sec:direction_vector_prediction} describes  direction vector generation and prediction, \sectionref{sec:teasar} the adaptation of TEASAR for vector-based skeletonization.

We formally represent vessel trees as directed acyclic graphs (DAG) \cite{diestel2017graph}:
Nodes correspond to 3D coordinates marking either branching points or sampled points along vessel segments, while edges denote the segments connecting them.
Each edge is assigned a radius characterizing the local vessel thickness. Edge directions follow a parent–child relationship, where the parent is the one closer to the vessel root.

\subsection{Direction Vector Generation and Prediction}
\label{sec:direction_vector_prediction}
We train a 3D U-Net~\cite{unet_ronneberger2015} to jointly predict a foreground mask and voxel-wise direction vectors (x, y, z components) from 3D grayscale images.
These vectors point toward the vessel centerline and are additionally biased rootward by a fixed stepsize, such that iterative following of the vectors eventually converges at the root.
Given a ground-truth foreground mask and a corresponding graph, we obtain training vectors as follows:
For each foreground voxel, we first identify the nearest edge in the ground-truth graph within the local vessel radius.
From the closest point on that edge, we then step a fixed distance toward the root, which defines the target point for the direction vector.
The resulting vectors point from each voxel to this upstream centerline point, yielding smaller magnitudes for voxels close to the centerline and larger ones near the vessel boundary.
Suppl.\ \figureref{fig:direction_vectors}a and \algorithmref{alg:vector_generation} provide details of the direction vectors and their generation.
During training, we use Binary Cross-Entropy (BCE) loss for the foreground mask and Mean Squared Error (MSE) loss for the direction vectors. 

\subsection{TEASAR-based Centerline Generation}
\label{sec:teasar}
TEASAR~\cite{sato2000teasar} extracts tree-shaped skeletons from volumetric tubular segmentation masks by placing centerlines in regions that lie maximally far from the object boundary.
It implements this by iteratively tracing shortest paths from a root to any voxel within the mask, applying a penalty that discourages paths from approaching the boundary.
This penalty depends solely on boundary distance, thus it may produce incorrect skeletons when vessels run in parallel, as shown in \figureref{fig:our_approach_summary}.
To address this, we propose a modified TEASAR variant that incorporates predicted voxel-wise direction vectors.
These vectors exhibit minimal magnitude and smallest angle relative to the centerline direction at voxels closest to the centerline.
We therefore augment the penalty term with components based on both \textit{vector magnitude} and \textit{angular deviation}. This additional penalty helps disambiguate vessels that are spatially tangent but semantically distinct.

\paragraph{Extended Penalty Term.}
\label{sec:teasar_flow_vectors}
Formally, let \( \Omega \subset \mathbb{R}^3 \) be the 3D object (foreground mask) and \( \partial \Omega \) be the object boundary (vessel surface).
Let \( p \in \Omega \) be the voxel inside the object located at the end of the shortest path $P$, constructed in a previous iteration step.
By \(N \subseteq \Omega \) we denote the set of all those adjacent voxels of \(p\) are lying within the object.
For each neighboring voxel $n\in N$, the \emph{distance from boundary field (DBF)} of \( n \), as leveraged by the original TEASAR, refers to the shortest Euclidean distance from voxel \( n \in N\) to the nearest boundary point \( b \in \partial \Omega \): 
\begin{equation}
    \text{DBF}(n) := \min_{b \in \partial \Omega} || n - b ||
\end{equation}
To incorporate directional guidance, we additionally consider 
the \emph{vector magnitude field (VMF)}, defined at each voxel $n$ as the magnitude of its direction vector $v_n$:  $\text{VMF}(n) := \left|| \vec{v}_n |\right|$. 
Note that $\text{VMF}(n)$ is minimal if the voxel lies on the centerline.
Furthermore, we denote by $\theta(p,n)\in[0,180]$
the angle (in degrees) between the direction vector $\vec{v}_p$ of \(p\) and the relative direction vector $\vec{r}:= n-p$ from $p$ to $n$ as shown in Suppl.\ \figureref{fig:direction_vectors}b.
Again $\theta(p,n)$ is minimal if $n$ is located in the direction of the predicted direction vector $\vec{v}_p$. 
The adapted penalty value we propose is given by:
\begin{equation}
\label{eq:pv_flow}
\text{PV}_{\text{flow}}(p,n) = 1{,}000{,}000 \cdot \Bigg( 
\left(1 - \frac{\text{DBF}(n)}{M_1}\right)^{16} + \\
\left(\frac{\text{VMF}(n)}{M_2}\right)^{16} + 
\left(\frac{\theta(p, n)}{M_3}\right)^{16} 
\Bigg)
\end{equation}
where 
\begin{equation}
    M_1 = \max_{p\in \Omega}(\text{DBF}(p))^{1.01},~ M_2 = \max_{p\in \Omega}(\text{VMF}(p))^{1.01},~ M_3 =180
\end{equation}

This directly adopts the original TEASAR penalty, augmenting it with VMF- and $\theta$- based terms of analogous form. Given this penalty, skeleton tracing proceeds as usual, starting from a most root-distant end point determined analogously as in original TEASAR, and appending the minimum-penalty neighboring node to the path until the root is reached.

\paragraph{Multi-root Processing and Adaptive Masking.}

\begin{figure}[t]
\centering
\includegraphics[width=\textwidth]{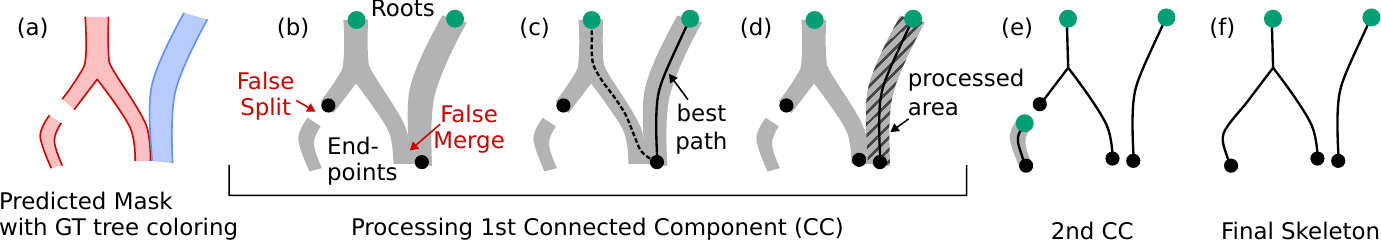}
\caption[Schematic of the modified TEASAR to address topological errors.]{
\label{fig:teasar}
\textbf{Addressing topological errors with the modified TEASAR algorithm.}
(a) Predicted foreground mask with distinct ground-truth (GT) trees shown in different colors.
(b) The algorithm selects one connected component and identifies its roots and endpoints.
(c) For each endpoint, paths are traced to all candidate roots, and the optimal path (with the lowest penalty) is chosen.
(d) Voxels within a specified radius around the traced path are marked as processed and excluded from subsequent tracing.
(e) After a component is fully processed, the algorithm proceeds to the next one; once all components are processed, disconnected fragments are evaluated for merging.
(f) Final output with complete centerlines.
}
\vspace{-15pt}
\end{figure}

Original TEASAR generates one skeleton per connected component of the binary mask.
However, segmentation errors may merge distinct vessels into a single component, producing structures with multiple trees and therefore multiple roots.
To address this, we extend TEASAR to support multiple roots within a component, as illustrated in \figureref{fig:teasar}.
Root locations in the datasets are provided either by manual annotation or automatically using the predicted direction vectors, effectively speeding up the annotation process.
Details on automated root detection are provided in Suppl.\ \sectionref{sec:automated_root_detection}.

Once the best path is established (\figureref{fig:teasar}(c) and (d)), original TEASAR applies a simple linear thresholding using a fixed scale and constant value $d = scale \cdot r + const$,
where $r$ is the vessel radius and $d$ is the masking distance used to exclude already processed regions. We extend this with an adaptive masking scheme in which both parameters vary smoothly with the local vessel radius.
This makes the method more robust across vessels of different radius and effectively suppresses spurious small branches in larger vessels.
Further details are provided in  Suppl.\ \sectionref{sec:adaptive masking}.

\paragraph{False Split Postprocessing.}
Finally, components without an assigned root are evaluated for potential merging with nearby trees. 
For each node in a disconnected component (the current node), we first identify neighboring nodes within a spatial distance of 5 voxels. Among these candidates, we retain only those that belong to a different tree and then compute their radius difference and angular deviation (based on the direction vectors) with respect to the current node. If the radius difference is below 3, the angular difference is below 100 degrees, and adding an edge between the two nodes does not introduce a cycle, we connect them. In cases with multiple valid neighbors, we select the closest one. This step helps to reduce false splits introduced by the segmentation. However, very small or isolated components that do not meet these criteria may remain disconnected.

\section{Evaluation}\label{sec:evaluation}

Our method predicts an acyclic vessel skeleton, with 3D coordinates assigned to each node, and edges oriented towards the root, forming a labeled directed acyclic graph (DAG). 
Comparing such graph against respective ground-truth (GT) is challenging \cite{Drees2019gerome,Lyu2022reta}: no standard metric exists (see Suppl. \ref{sec:metrics_discussion}), and many measures lack intuitive topological meaning or depend sensitively on node matching and sampling.
To compare predicted and GT graphs, we first resample both at a fixed step size $s>0$.
The next essential step is an \emph{assignment strategy} (cf.\ \citet{metrics_reloaded2024}) that matches nodes and edges between predicted and GT graphs. 
In Sec.\ \ref{subsec:hierarchical_matching} we propose a greedy hierarchical matching procedure designed for robust topological correspondence.
Based on these correspondences, we compute error metrics as described in Sec.\ \ref{sec:metrics_def}.

\subsection{Hierarchical Matching}
\label{subsec:hierarchical_matching}
Commonly used approaches to assign nodes or edges of two graphs to each other, is greedy nearest-neighbor or optimal matching based solely on spatial proximity, such as in \citet{Drees2019gerome,trexplorer_super2025}.
While effective in simple scenarios, this strategy ignores the structural and semantic information inherent to tree-like graphs, making it unsuitable for capturing topological similarity—particularly in cases where different vessels are close-by.
Therefore, we propose a greedy one-to-at-most-one hierarchical matching scheme that incorporates spatial, semantic and ancestor information.
It is similar to \citet{Gillette2011DIAMEMetric}, but also applicable in multi-tree scenarios.

A pseudo-code description of the matching procedure is provided in Suppl.~\algorithmref{alg:hierarchical-matching}. In short, first, the connected components of the GT graph $G$ and predicted graph $P$ are determined. 
Each node in $G$ and $P$ is assigned a semantic class---\textit{root}, \textit{branching point}, \textit{leaf}, or \textit{intermediate}. 
For every node in $G$, we identify candidate nearest neighbors in $P$ within a predefined distance threshold and rank them, first by semantic correspondence and second by spatial proximity.
We then iterate over the GT roots, always selecting the next root whose best candidate exhibits the highest matching priority (i.e., first by identical semantic class and second by minimal distance).
Starting from each root, we perform \textit{two depth-first traversals}.
In the first, we visit branching and leaf nodes, and assign them to the best available candidate based on the matching status of the candidate’s parent, the candidate's semantic label, and its distance. 
Thus, candidates whose parents are matched within the same GT tree receive highest priority.
If no suitable candidate exists, the GT node remains unmatched.
The second traversal processes intermediate nodes using the same criteria.
After completing a GT tree, we proceed to the next. 
Importantly, the candidate lists are updated immediately whenever two nodes become matched to maintain consistency throughout the hierarchy. 
A quantitative comparison with greedy nearest-neighbor and Hungarian matching is presented in Suppl.\ \tableref{tab:matching_comparison}.

\subsection{Metrics Definitions}\label{sec:metrics_def}
\begin{figure}
    \centering
    \scriptsize
    \resizebox{1.0\linewidth}{!}{\includegraphics{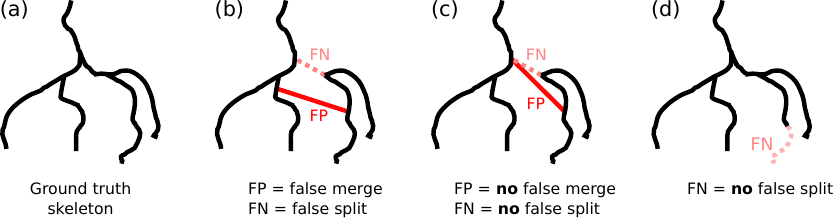}}
    \caption{\textbf{False Merges \& False Splits.}
    (a) A ground-truth skeleton next to three possible predictions.
    (b) The predicted graph has one FN and one FP edge. The FP is a false merge since it connects two nodes which are not ancestor of each other. Consequently, the FN is a false split. 
    (c) The FP edge is \emph{not} a false merge since it keeps the ancestor relation w.r.t. to its parent node intact. Consequently, the FN is \emph{not} a false split.
    (d) The predicted graph only has FN edges which are \emph{not} false splits since they do not change the node ancestor relations.
    \label{fig:fm_fs}}
\end{figure}

Based on our literature review in Suppl.\ \sectionref{sec:metrics_discussion}, we report the \emph{edge-wise} F1 score, as used in \citet{Drees2019gerome,Drees2021voreenSkel}, since the F1 score is a widely established and, in our view, easily interpretable measure of topological correctness when applied to edges.
Yet F1 alone does not capture the structural impact of certain errors. For instance, a false positive edge connecting unrelated nodes can distort the topology far more than a shortcut to an ancestor (see \figureref{fig:fm_fs}).
To address this, we introduce \textit{false splits} and \textit{false merges} as additional topology-aware error measures, extending prior work \cite{matula2015tra,fisbe} to multi-tree graphs where errors may arise both within and across trees.
In addition, we report the metrics used in \citet{trexplorer_super2025} for comparability reasons.
Although they also report F1 scores at the node and branch level—similar in spirit to our recommendation—their metrics rely on greedy one-to-one nearest-neighbor matching and the computation operates on individual nodes, thereby not fully capturing connectivity.
Regarding graph-level Betti numbers, only Betti–0 (the number of connected components) is meaningful; because assuming only trees, Betti–1 (the number of cycles) is always zero.

\paragraph{Edge-wise F1 Score.}
In the following, we denote by $G$ and $P$ the GT and predicted graphs with node sets $V_G$ and $V_P$, and by $\Phi: V_G \rightarrow V_P$ the one-to-one node matching.
Both graphs are resampled to a fixed step size ($s=1$ voxel, unless stated otherwise), after which we apply our proposed hierarchical matching.
The edge-wise F1 score is computed as the balanced measure of precision and recall, relating the number of correctly matched edges (true positives, TP) in $G$ to the number of incorrectly matched (false positives, FP) or incorrectly unmatched (false negatives, FN) edges, where TP, FP and FN are defined as:
\begin{itemize}
    \item An edge \( (v,v') \) in \( G \) is a TP if and only if $(\Phi(v), \Phi(v'))$ is an edge in $P$.
    \item An edge \( (v,v') \) in \( G \) is a FN if and only if $v$ and $v'$ were matched and $(\Phi(v), \Phi(v'))$ is \textit{not} an edge in $P$.
    \item An edge \( (\Phi(v),\Phi(v')) \) in \( P \) is a FP if and only if $(v,v')$ is \textit{not} an edge in $G$.
\end{itemize}

The \emph{edge-wise F1, precision and recall} are defined as
\[
F_1^{\text{edge}} := \frac{2\text{TP}}{2\text{TP} + \text{FP} + \text{FN}}, \quad
Precision^{\text{edge}} := \frac{TP}{TP+FP}, \quad Recall^{\text{edge}} := \frac{TP}{TP+FN}
\]

\paragraph{False Merges (FM) and False Splits (FS).}
We further define false merges and false splits as follows.
A false merge is a FP edge \( (\Phi(v), \Phi(v')) \) in $P$ where the GT nodes \( v \) and \( v' \) in $G$ have no directed path between them. In other words, neither is an ancestor of the other. 
For a false split, we consider the subgraph \( P^* \) of \( P \) obtained by excluding all false merge edges. A false split is a FN edge \( (v,v') \) in \( G \) such that adding the missing edge \( (\Phi(v), \Phi(v')) \) to \( P^* \) merges two connected components into one. This means FS correspond to missing edges in \( P \) that cause false disconnections in \( P^* \). 
The number of FS can be determined by
$\beta_0(P^*) - \beta_0(G)$, since each FS increases the number of connected components of $P^*$. Here, $\beta_0$ is the number of connected components (Betti-0).

\paragraph{Trexplorer-super Evaluation for Comparability.}
To compute the metrics from \citet{trexplorer_super2025}, both $G$ and $P$ are resampled to one voxel spacing.
At the node level, precision, recall, and F1 are reported, along with radius accuracy measured via the mean absolute error (MAE).
At the branch level, the F1 score is reported; where a branch is considered a TP if at least 80\% of its nodes are matched.

\section{Experiments}\label{sec:experiments}
We evaluate our method on four vascular datasets, covering both single and multi-tree scenarios.
Consistent with prior works \cite{trexplorer_super2025, trexplorer2024, vesselformer2024}, all reported experiments use ground-truth root locations as input to our adapted TEASAR algorithm. Dataset-specific modifications of the training procedure, together with details on sample sizes and the data splits, are in Suppl. \sectionref{sec:ext_experiments}. An ablation study quantifying the contributions of each component of our method (Suppl. \sectionref{sec:ablation_study}), a vector noise sensitivity study (Suppl. \sectionref{sec:vector noise sensitivity}) and test-time noise sensitivity (Suppl. \sectionref{sec:gaussian noise sensitivity}) are also included. We also evaluate the effect of replacing the U-Net in our method with nnU-Net~\citep{isensee2021nnu} in Suppl.~\sectionref{sec:nnunet}.

\subsection{Model Architecture and Training}
For segmentation and vector prediction, we employ a 4-layer U-Net~\cite{unet_ronneberger2015} with batch normalization and 16 initial feature channels, which double at each downsampling step.
The network is trained on randomly sampled input patches.
Data augmentation includes intensity shifts and randomly masking out $3\times3\times3$ voxel crops.
Training is performed for 300{,}000 iterations with a batch size of 1 using the Adam optimizer.
Aside from dataset-specific input sizes and augmentations, the architecture and training protocol are kept identical across all datasets. We have mentioned the further training details and different settings in Suppl.~\sectionref{sec:training_settings}.

\subsection{Case 1: Single-Tree Data}
For the single-tree datasets, we compare our method against Vesselformer \cite{vesselformer2024}, Trexplorer \cite{trexplorer2024}, and Trexplorer-super \cite{trexplorer_super2025}.
We report all baselines and metrics as in \cite{trexplorer_super2025}, as we were unable to reproduce their published results and thus cannot faithfully compare in terms of our new metrics.
Instead, we follow their evaluation protocol to enable a fair comparison.
Thus we report point-level F1, precision, recall, and radius MAE, as well as branch-level F1 and Betti scores in \tableref{tab:point_metrics}.
Apart from that, we also evaluated the single-tree datasets using our own metrics in \tableref{tab:single tree our metrics} to support future benchmarking.

The \textbf{Single-Tree Synthetic} dataset, introduced in \citet{trexplorer_super2025}, is generated using the Synthetic Vascular Toolkit (SVT) \cite{sexton2025svt}.
Each volume contains a single vascular tree, its segmentation mask, and the corresponding 3D centerline graph.
As shown in \tableref{tab:point_metrics}, our model consistently outperforms the current state-of-the-art across both point-level and branch-level metrics.
Suppl.\ \figureref{fig:qualitative_syn_st} shows qualitative results.

The publicly available \textbf{Parse 2022} pulmonary artery segmentation dataset \cite{luo2024parse} contains 100 computed tomography pulmonary angiography (CTPA) volumes with pixel-wise segmentation masks.
These masks were created semi-automatically by experts using a region-growing approach.
\citet{trexplorer_super2025} subsequently derived centerlines from these masks using the Kimimaro TEASAR implementation \cite{Silversmith_Kimimaro_Skeletonize_densely_2021}.
Note that these ground-truth centerlines were generated \textit{automatically}.
As shown in \tableref{tab:point_metrics}, our model outperforms the current state-of-the-art on both point-level and branch-level F1.
However, at graph level we note some Betti-0 errors: although our post-processing step is designed to correct false splits, it does not fully guarantee global connectivity of the predicted vascular tree.
Suppl.\ \figureref{fig:qualitative_parse} shows qualitative results. We note that a potential bias may favor our method, since the ground-truth skeletons are generated using the TEASAR algorithm.

\begin{table}
\centering
\scriptsize
\setlength{\tabcolsep}{4pt}
\caption{Quantitative comparison of our method with Vesselformer, Trexplorer and Trexplorer Super for the Single-Tree Synthetic and Parse 2022 datasets.
Please note that we report all baselines and metrics as in \citet{trexplorer_super2025}.
Our results are reported as mean and standard deviation (±) over three independent runs, while baseline results are reported over five runs.
}
\begin{tabular}{clcccccccc}
\toprule
& \multirow{2}{*}{Model} 
& \multicolumn{4}{c}{Point Level} 
& \multicolumn{1}{c}{Branch Level}
& \multicolumn{2}{c}{Graph Level}\\
\cmidrule(lr){3-6} \cmidrule(lr){7-7}\cmidrule(lr){8-9}
& & F1$\uparrow$ & Prec$\uparrow$ & Rec$\uparrow$ & Rad.(MAE)$\downarrow$ 
 & F1$\uparrow$ & $\beta_0\downarrow$ & $\beta_1\downarrow$\\
\midrule
\parbox[t]{1mm}{\multirow{4}{*}{\rotatebox[origin=c]{90}{\textbf{\tiny Synthetic}}}}
& Vesselformer & $48.18${\tiny $\pm5.62$} & $44.53${\tiny $\pm7.87$} & $61.52${\tiny $\pm1.14$} & $0.42${\tiny $\pm0.01$} & $15.95${\tiny $\pm0.36$}& $81.7${\tiny $\pm16.8$} & $653.5${\tiny $\pm138.7$}\\
& Trexplorer & $39.40${\tiny $\pm8.62$} & $30.91${\tiny $\pm9.45$} & $78.21${\tiny $\pm4.13$} & $0.23${\tiny $\pm0.03$} & $26.26${\tiny $\pm7.18$} & $0${\tiny $\pm0.0$} & $0${\tiny $\pm0.0$}\\
& Trexpl. Super & $77.83${\tiny $\pm1.89$} & $91.91${\tiny $\pm3.28$} & $70.44${\tiny $\pm3.02$} & $\mathbf{0.1}${\tiny $\pm0.01$} & $77.12${\tiny $\pm1.59$} & $0${\tiny $\pm0.0$} & $0${\tiny $\pm0.0$}\\
& \textbf{Ours} & $\mathbf{92.25}${\tiny $\pm0.02$} & $\mathbf{95.49}${\tiny $\pm0.01$} & $\mathbf{89.24}${\tiny $\pm0.05$} & $0.29${\tiny $\pm0.00$} & $\mathbf{81.50}${\tiny $\pm0.16$} & $0${\tiny $\pm0.0$} & $0${\tiny $\pm0.0$}\\
\midrule
\parbox[t]{1mm}{\multirow{4}{*}{\rotatebox[origin=c]{90}{\textbf{\tiny Parse2022}}}}
& Vesselformer & $16.43${\tiny $\pm0.78$} & $18.49${\tiny $\pm1.84$} & $15.28${\tiny $\pm0.83$} & $1.11${\tiny $\pm0.03$} & $1.99${\tiny $\pm0.16$} & $410${\tiny $\pm23.9$} & $246.7${\tiny $\pm78.1$}\\
& Trexplorer & $10.01${\tiny $\pm4.98$} & $9.87${\tiny $\pm3.76$} & $12.01${\tiny $\pm7.46$} & $1.21${\tiny $\pm0.30$} & $3.71${\tiny $\pm1.91$} & $0${\tiny $\pm0.0$} & $0${\tiny $\pm0.0$}\\
& Trexpl. Super & $39.46${\tiny $\pm1.93$} & $55.27${\tiny $\pm3.00$} & $33.99${\tiny $\pm3.34$} & $\mathbf{0.56}${\tiny $\pm0.01$} & $23.46${\tiny $\pm1.09$} & $0${\tiny $\pm0.0$} & $0${\tiny $\pm0.0$}\\
& \textbf{Ours} & $\mathbf{57.52}${\tiny $\pm0.66$} & $\mathbf{59.11}${\tiny $\pm0.37$} & $\mathbf{57.81}${\tiny $\pm0.89$} & $0.58${\tiny $\pm0.02$} & $\mathbf{35.33}${\tiny $\pm1.19$} & $1.85${\tiny $\pm0.46$} & $0${\tiny $\pm0.0$}\\

\bottomrule
\end{tabular}
\label{tab:point_metrics}
\end{table}

\begin{table}
    \centering
    \scriptsize
    \caption{\label{tab:single tree our metrics}Quantitative results of our method on the Single-Tree datasets using our proposed evaluation metrics to support future benchmarking.
    We report mean and standard deviation (±) over three independent runs.
    }
    \begin{tblr}{width=\linewidth, rows={abovesep=1pt, belowsep=1pt},
    } 
        \midrule
        \SetCell[r=2]{l}{Dataset}
            & \SetCell[c=3]{c}{Edges} & & & 
            \SetCell[c=2]{c}{FM$\downarrow$} & & \SetCell[c=2]{c}{FS$\downarrow$} &\\
        \cmidrule[lr]{2-4} \cmidrule[lr]{5-6} \cmidrule[lr]{7-8}
         & F1$\uparrow$ & Prec$\uparrow$ & Rec$\uparrow$
            & Rel. & Abs. & Rel. & Abs. \\
        \midrule
        Synthetic & 
          $0.89${\tiny $\pm0.001$} & $0.93${\tiny $\pm0.001$} & $0.87${\tiny $\pm0.001$} & 
          $0.010${\tiny $\pm0.0$} & $25.86${\tiny $\pm0.10$}  & $0.009${\tiny $\pm0.0$}  & $25.86${\tiny $\pm0.10$}
           \\
        Parse2022 &
          $0.69${\tiny $\pm0.015$}  & $0.90${\tiny $\pm0.004$} & $0.57${\tiny $\pm0.012$} &  
          $0.007${\tiny $\pm0.0$} & $83.8${\tiny $\pm1.37$}  & $0.004${\tiny $\pm0.0$}  & $85.62${\tiny $\pm1.68$}  \\
        \bottomrule
    \end{tblr}
\end{table}

\subsection{Case 2: Multi-Tree Data}
Although vascular networks are ideally single-tree structures, real data often contain multiple trees due to challenges in separating arteries and veins or imaging artifacts.
Here, we report our recommended metrics, namely edge-level F1, precision, recall, false merges (FM), and false splits (FS).
We compare Vesselpose against different segmentation-based approaches, where we skeletonize the resulting binary masks with Kimimaro TEASAR.

\begin{figure}
  \centering

  \makebox[0.9\linewidth]{%
    \subfigure[Ground-truth]{%
      \includegraphics[width=0.30\linewidth]{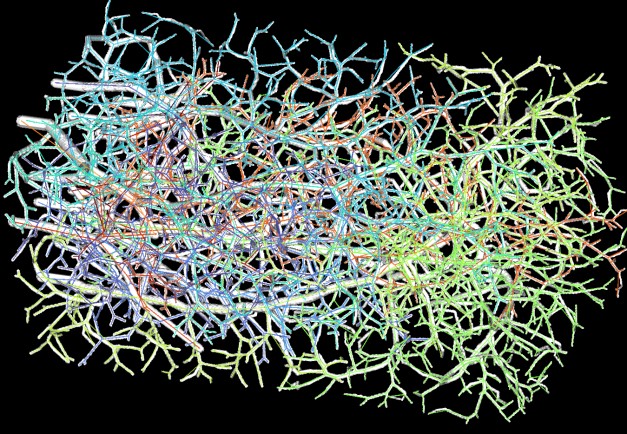}
    }\hfill
    \subfigure[Ours]{%
      \includegraphics[width=0.30\linewidth]{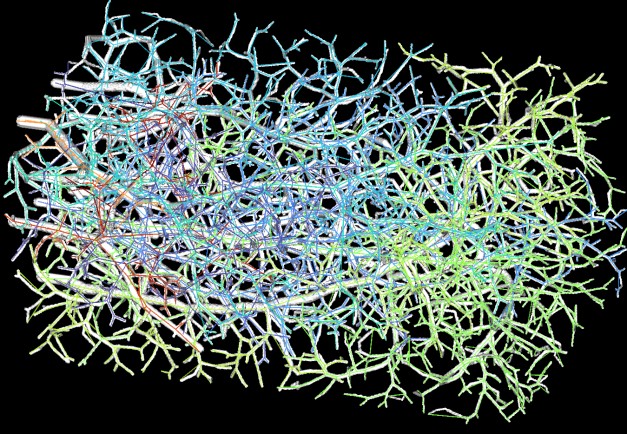}
    }\hfill
    \subfigure[U-Net + TEASAR]{%
      \includegraphics[width=0.30\linewidth]{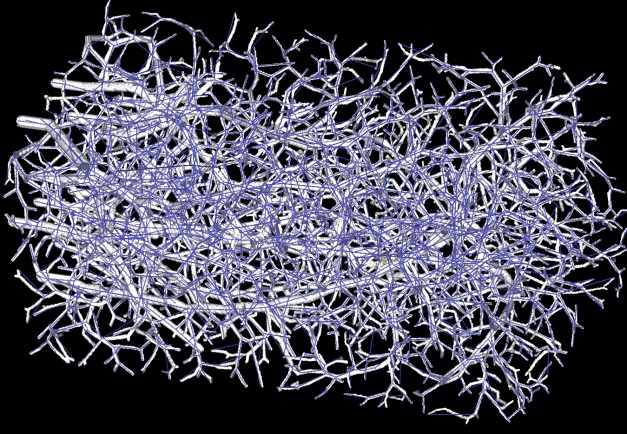}
    }%
  }

  \vspace{4pt}

  \makebox[0.9\linewidth]{%
    \subfigure[Ground-truth]{%
      \includegraphics[width=0.45\linewidth]{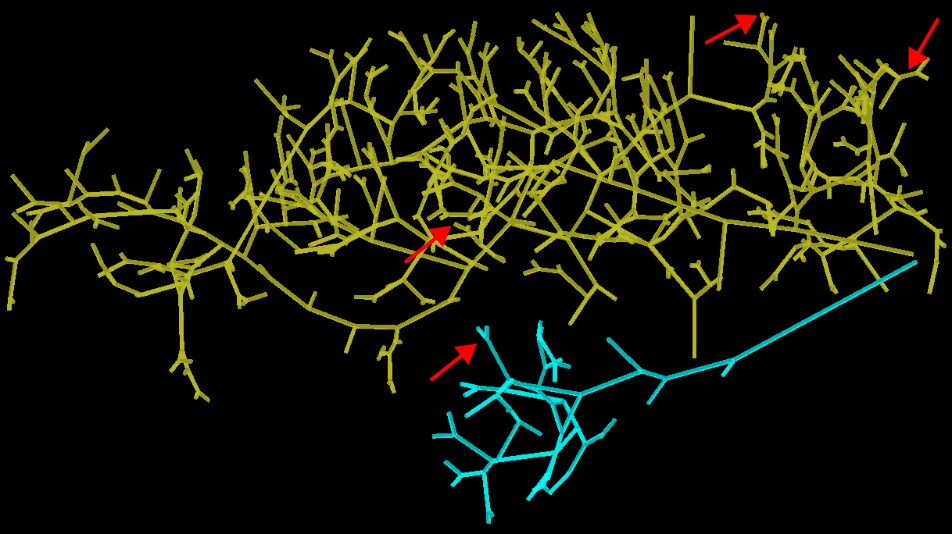}
    }\hfill
    \subfigure[Ours]{%
      \includegraphics[width=0.45\linewidth]{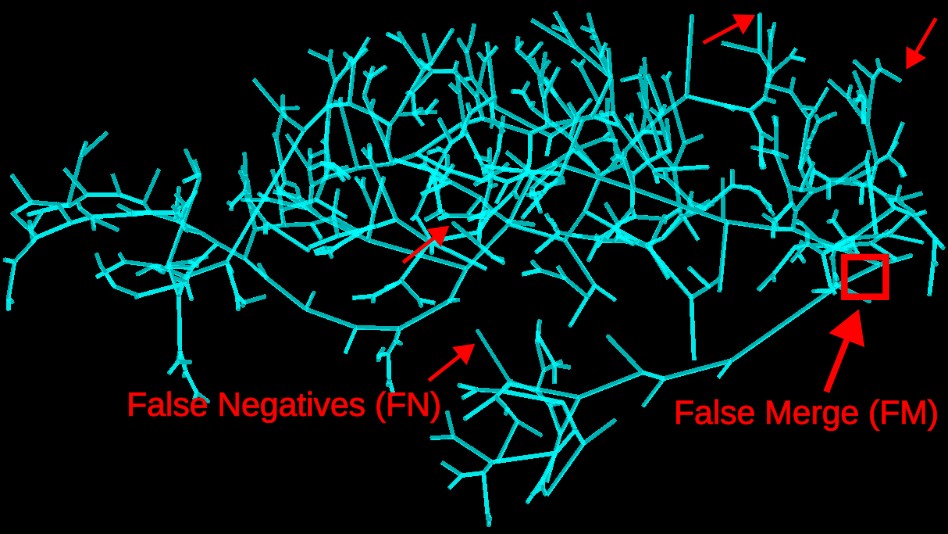}
    }%
  }

  \caption{\textbf{Qualitative results for the multi-tree synthetic dataset.}
  First row: Segmentation mask and skeletons overlaid, where each color represents a distinct tree. 
  Our approach separates most trees, whereas U-Net + TEASAR merge all trees into one component.
  Second row: Failure cases for our method, including missed small terminal branches (red arrows) 
  and falsely merged trees (red rectangle).}
  \label{fig:qualitative_multitree_synthetic}
\end{figure}

The \textbf{Multi-Tree Synthetic} dataset originates from \citet{tetteh2019_deepvesselnet} and is generated using vessel formation simulations \cite{schneider2012tissue}.
We compare our method to vesselFM \cite{Wittmann_2025_CVPR} and a standard U-Net \cite{unet_ronneberger2015}.
We also report an upper bound by applying TEASAR directly to the ground-truth masks of \citet{schneider2012tissue}.
The results in \tableref{tab:multi_tree_comp} show that our method consistently outperforms these segmentation-based baselines.
Original TEASAR produces one tree per connected component, but baseline segmentations frequently merge distinct trees into a single component.
As a result, their false merge rates are substantially higher than our method.
In \figureref{fig:qualitative_multitree_synthetic} we show qualitative results and discuss failure cases of our method.

The \textbf{Multi-Tree Micro-CT} data were acquired from perfused rat hearts using a solidifying Microfil contrast agent, using a protocol broadly similar to that described in \cite{napieczynska2024muct}. This approach provides strong vascular contrast and enables visualization of small vessels.
The dataset is still under study and may be made publicly available at a later stage.
We use four rat heart volumes: one to fine-tune a U-Net pretrained on the synthetic multi-tree data, and three for validation and testing. For these, we annotated three $400 \times 400 \times 400$ voxel crops using CATMAID \cite{catmaid2009,catmaid2016}.
Details on the data, model, and fine-tuning procedure are provided in Suppl.\ \sectionref{sec:ext_micro_ct}.
\tableref{tab:multi_tree_comp} shows that our method consistently outperforms a standard U-Net with TEASAR.
Although the dataset is relatively small, the observed performance improvement is consistent with those reported on the other datasets.
Our higher absolute FM and FS values stem from reconstructing more complete skeletons, whereas U-Net and regular TEASAR miss large graph regions—reflected in our correspondingly lower relative FM/FS counts, also seen in Suppl. \figureref{fig:qualitative_micro-ct}.
 
\begin{table}
    \centering
    \scriptsize
    \caption{\label{tab:multi_tree_comp}Quantitative comparison of our method on the Multi-Tree Synthetic and Micro-CT Heart datasets.
    We compare against U-Net, vesselFM, and the ground-truth (GT) segmentation, each followed by TEASAR skeletonization.
    Because a more complete prediction can yield higher absolute FM and FS counts than an incomplete graph, we additionally report relative FM/FS values, obtained by dividing the absolute counts by the total number of predicted edges.
    We report mean and standard deviation (±) over three independent runs (except for vesselFM and GT).
    }
    \begin{tblr}{width=\linewidth, rows={abovesep=1pt, belowsep=1pt},
    cell{3}{1} = {r=4}{cmd=\rotatebox{90}},
    cell{7}{1} = {r=2}{cmd=\rotatebox{90}},
    } 
        \midrule
        & \SetCell[r=2]{l}{Method}  
            & \SetCell[c=3]{c}{Edges} & & & 
            \SetCell[c=2]{c}{FM$\downarrow$} & & \SetCell[c=2]{c}{FS$\downarrow$} &\\
        \cmidrule[lr]{3-5} \cmidrule[lr]{6-7} \cmidrule[lr]{8-9}
            & & F1$\uparrow$ & Prec$\uparrow$ & Rec$\uparrow$
            & Rel. & Abs. & Rel. & Abs. \\
        \midrule
        \textbf{\makecell{\tiny Multi-Tree\\\tiny Synthetic}} & U-Net 
          & $0.46${\tiny $\pm0.001$} & $0.64${\tiny $\pm0.002$} & $0.36${\tiny $\pm0.001$} 
          & $0.02${\tiny $\pm0$} & $51.87${\tiny $\pm0.68$} 
          & $0.01${\tiny $\pm0.002$} & $51.70${\tiny $\pm3.39$} \\
        & VesselFM 
            & 0.46 & 0.62 & 0.36 
            & 0.02 & 51.3 
            & 0.01 & 58.3 \\
        & GT Segm 
            & 0.46 & 0.64 & 0.36 
            & 0.02 & 52.1 
            & 0.01 & 38.4 \\
        & Ours
            & $\mathbf{0.80}${\tiny $\pm0.002$} & $\mathbf{0.79}${\tiny $\pm0.002$} & $\mathbf{0.80}${\tiny $\pm0.001$}
            & $\mathbf{0.007}${\tiny $\pm0$} & $\mathbf{30.80}${\tiny $\pm1.10$}
            & $\mathbf{0.007}${\tiny $\pm0$} & $\mathbf{29.67}${\tiny $\pm1.80$} \\
        \midrule
        \textbf{\makecell{\tiny Micro\\\tiny CT}} & U-Net 
            & $0.32${\tiny $\pm0.03$} & $0.22${\tiny $\pm0.04$} & $0.57${\tiny $\pm0.01$} 
            & $0.01${\tiny $\pm0$} & $\mathbf{26.25}${\tiny $\pm2.25$} 
            & $0.009${\tiny $\pm0$} & $\mathbf{23.5}${\tiny $\pm4.2$} \\
        & Ours 
            & $\mathbf{0.50}${\tiny $\pm0.002$} & $\mathbf{0.43}${\tiny $\pm0.001$} & $\mathbf{0.63}${\tiny $\pm0.002$}
            & $\mathbf{0.006}${\tiny $\pm0$} & $45.5${\tiny $\pm1.4$}
            & $\mathbf{0.006}${\tiny $\pm0$} & $42.5${\tiny $\pm1.4$} \\
        \bottomrule
    \end{tblr}
\end{table}

\section{Conclusion}
\label{sec:conclusion}
In this work, we presented a novel method for extracting vessel graphs from 3D images.
We demonstrated its effectiveness on four datasets, comprising synthetic and real data, as well as single-tree and multi-tree vascular structures.
In addition, we introduced a hierarchical graph matching algorithm that yields more topologically meaningful node and edge correspondences, and we defined false splits and false merges as intuitive topology-aware error measures for tree graphs.
Despite these advances, several evaluation metrics such as the edge-wise F1 score remain sensitive to different node sampling and matching strategies.
A more systematic analysis of these effects represents a valuable direction for future work.
Moreover, there is a pressing need for publicly available, real-world datasets with high-quality, manually annotated 3D vessel graphs.
Such datasets, combined with an established evaluation protocol encompassing sampling, matching, and metrics, would be essential for enabling consistent benchmarking and further methodological development in this area.

\clearpage  
\midlacknowledgments{This work was supported by the Berlin Institute of Health (BIH) Research Focus Area Vascular Biomedicine Grant (K11000200102/3), 
German Centre for Cardiovascular Research (DZHK) - Excellence Programme Postdoc Start-up Grant (81X3100109) for animal experiments, and German Research Foundation (DFG) Individual Research Grant UMDISTO (project no.\ 498181230). We thank Hanna Napieczy\'nska and the Animal Phenotyping Platform of the Max Delbr\"uck center for kindly providing unpublished rat heart data.   We thank the Kainmueller Lab for their support and feedback.}

\bibliography{midl26_101}

@String(CVPR= {IEEE Conf. Comput. Vis. Pattern Recog.})

@String(ECCV= {Eur. Conf. Comput. Vis.})

@String(CVPR  = {CVPR})

@String(ECCV  = {ECCV})

@misc{tetteh2019_deepvesselnet,
      title={DeepVesselNet: Vessel Segmentation, Centerline Prediction, and Bifurcation Detection in 3-D Angiographic Volumes}, 
      author={Giles Tetteh and Velizar Efremov and Nils D. Forkert and Matthias Schneider and Jan Kirschke and Bruno Weber and Claus Zimmer and Marie Piraud and Bjoern H. Menze},
      year={2019},
      eprint={1803.09340},
      archivePrefix={arXiv},
      primaryClass={cs.CV},
      url={https://arxiv.org/abs/1803.09340}, 
}

@INPROCEEDINGS{sato2000teasar,
  author={Sato, M. and Bitter, I. and Bender, M.A. and Kaufman, A.E. and Nakajima, M.},
  booktitle={Proceedings the Eighth Pacific Conference on Computer Graphics and Applications}, 
  title={TEASAR: tree-structure extraction algorithm for accurate and robust skeletons}, 
  year={2000},
  volume={},
  number={},
  pages={281-449},
  doi={10.1109/PCCGA.2000.883951}}

@InProceedings{trexplorer2024,
        author = { Naeem, Roman and Hagerman, David and Svensson, Lennart and Kahl, Fredrik},
        title = { { Trexplorer: Recurrent DETR for Topologically Correct Tree Centerline Tracking } },
        booktitle = {proceedings of Medical Image Computing and Computer Assisted Intervention -- MICCAI 2024},
        year = {2024},
        publisher = {Springer Nature Switzerland},
        volume = {LNCS 15011},
        month = {October},
        page = {744 -- 754}
}

@InProceedings{trexplorer_super2025,
author="Naeem, Roman
and Hagerman, David
and Alv{\'e}n, Jennifer
and Svensson, Lennart
and Kahl, Fredrik",
editor="Gee, James C.
and Alexander, Daniel C.
and Hong, Jaesung
and Iglesias, Juan Eugenio
and Sudre, Carole H.
and Venkataraman, Archana
and Golland, Polina
and Kim, Jong Hyo
and Park, Jinah",
title="Trexplorer Super: Topologically Correct Centerline Tree Tracking of Tubular Objects in CT Volumes",
booktitle="Medical Image Computing and Computer Assisted Intervention -- MICCAI 2025",
year="2025",
publisher="Springer Nature Switzerland",
address="Cham",
pages="595--605",
isbn="978-3-032-04984-1"
}

@InProceedings{vesselformer2024,
  title = 	 {Vesselformer: Towards Complete 3D Vessel Graph Generation from Images},
  author =       {Prabhakar, Chinmay and Shit, Suprosanna and Paetzold, Johannes C. and Ezhov, Ivan and Koner, Rajat and Li, Hongwei and Kofler, Florian Sebastian and Menze, Bjoern},
  booktitle = 	 {Medical Imaging with Deep Learning},
  pages = 	 {320--331},
  year = 	 {2024},
  editor = 	 {Oguz, Ipek and Noble, Jack and Li, Xiaoxiao and Styner, Martin and Baumgartner, Christian and Rusu, Mirabela and Heinmann, Tobias and Kontos, Despina and Landman, Bennett and Dawant, Benoit},
  volume = 	 {227},
  series = 	 {Proceedings of Machine Learning Research},
  month = 	 {10--12 Jul},
  publisher =    {PMLR},
  url = 	 {https://proceedings.mlr.press/v227/prabhakar24a.html}
}

@article{cellpose2021,
author={Stringer, Carsen
and Wang, Tim
and Michaelos, Michalis
and Pachitariu, Marius},
title={Cellpose: a generalist algorithm for cellular segmentation},
journal={Nature Methods},
year={2021},
month={Jan},
day={01},
volume={18},
number={1},
pages={100-106},
issn={1548-7105},
doi={10.1038/s41592-020-01018-x},
url={https://doi.org/10.1038/s41592-020-01018-x}
}

@article{cellpose2022,
author={Pachitariu, Marius
and Stringer, Carsen},
title={Cellpose 2.0: how to train your own model},
journal={Nature Methods},
year={2022},
month={Dec},
day={01},
volume={19},
number={12},
pages={1634-1641},
issn={1548-7105},
doi={10.1038/s41592-022-01663-4},
url={https://doi.org/10.1038/s41592-022-01663-4}
}

@article{voreen2009,
  title={Voreen: A Rapid-Prototyping Environment for Ray-Casting-Based Volume Visualizations},
  author={Jennis Meyer-Spradow and Timo Ropinski and J{\"o}rg Mensmann and Klaus H. Hinrichs},
  journal={IEEE Computer Graphics and Applications},
  year={2009},
  volume={29},
  pages={6-13},
  url={https://api.semanticscholar.org/CorpusID:8211514}
}

@article{metrics_reloaded2024,
author={Maier-Hein, Lena
and Reinke, Annika
and Godau, Patrick
and Tizabi, Minu D.
and Buettner, Florian
and Christodoulou, Evangelia
and Glocker, Ben
and Isensee, Fabian
and Kleesiek, Jens
and Kozubek, Michal
and Reyes, Mauricio
and Riegler, Michael A.
and Wiesenfarth, Manuel
and Kavur, A. Emre
and Sudre, Carole H.
and Baumgartner, Michael
and Eisenmann, Matthias
and Heckmann-N{\"o}tzel, Doreen
and R{\"a}dsch, Tim
and Acion, Laura
and Antonelli, Michela
and Arbel, Tal
and Bakas, Spyridon
and Benis, Arriel
and Blaschko, Matthew B.
and Cardoso, M. Jorge
and Cheplygina, Veronika
and Cimini, Beth A.
and Collins, Gary S.
and Farahani, Keyvan
and Ferrer, Luciana
and Galdran, Adrian
and van Ginneken, Bram
and Haase, Robert
and Hashimoto, Daniel A.
and Hoffman, Michael M.
and Huisman, Merel
and Jannin, Pierre
and Kahn, Charles E.
and Kainmueller, Dagmar
and Kainz, Bernhard
and Karargyris, Alexandros
and Karthikesalingam, Alan
and Kofler, Florian
and Kopp-Schneider, Annette
and Kreshuk, Anna
and Kurc, Tahsin
and Landman, Bennett A.
and Litjens, Geert
and Madani, Amin
and Maier-Hein, Klaus
and Martel, Anne L.
and Mattson, Peter
and Meijering, Erik
and Menze, Bjoern
and Moons, Karel G. M.
and M{\"u}ller, Henning
and Nichyporuk, Brennan
and Nickel, Felix
and Petersen, Jens
and Rajpoot, Nasir
and Rieke, Nicola
and Saez-Rodriguez, Julio
and S{\'a}nchez, Clara I.
and Shetty, Shravya
and van Smeden, Maarten
and Summers, Ronald M.
and Taha, Abdel A.
and Tiulpin, Aleksei
and Tsaftaris, Sotirios A.
and Van Calster, Ben
and Varoquaux, Ga{\"e}l
and J{\"a}ger, Paul F.},
title={Metrics reloaded: recommendations for image analysis validation},
journal={Nature Methods},
year={2024},
month={Feb},
day={01},
volume={21},
number={2},
pages={195-212},
issn={1548-7105},
doi={10.1038/s41592-023-02151-z},
url={https://doi.org/10.1038/s41592-023-02151-z}
}

@InProceedings{unet_ronneberger2015,
author="Ronneberger, Olaf
and Fischer, Philipp
and Brox, Thomas",
editor="Navab, Nassir
and Hornegger, Joachim
and Wells, William M.
and Frangi, Alejandro F.",
title="U-Net: Convolutional Networks for Biomedical Image Segmentation",
booktitle="Medical Image Computing and Computer-Assisted Intervention -- MICCAI 2015",
year="2015",
publisher="Springer International Publishing",
address="Cham",
pages="234--241",
isbn="978-3-319-24574-4"
}

@article{napieczynska2024muct,
  title={$\mu$CT imaging of a multi-organ vascular fingerprint in rats},
  author={Napieczy{\'n}ska, Hanna and Kedziora, Sarah M and Haase, Nadine and M{\"u}ller, Dominik N and Heuser, Arnd and Dechend, Ralf and Kr{\"a}ker, Kristin},
  journal={Plos one},
  volume={19},
  number={10},
  pages={e0308601},
  year={2024},
  publisher={Public Library of Science San Francisco, CA USA}
}

@article{Gillette2011DIAMEMetric,
  title        = {The DIADEM Metric: Comparing Multiple Reconstructions of the Same Neuron},
  author       = {Gillette, Todd A. and Brown, Keith M. and Ascoli, Giorgio A.},
  journal      = {Neuroinformatics},
  volume       = {9},
  number       = {2-3},
  pages        = {233--245},
  year         = {2011},
  publisher    = {Springer},
  doi          = {10.1007/s12021-011-9117-y}
}

@inproceedings{rempfler2016minimum,
  title={The minimum cost connected subgraph problem in medical image analysis},
  author={Rempfler, Markus and Andres, Bjoern and Menze, Bjoern H},
  booktitle={Medical Image Computing and Computer-Assisted Intervention-MICCAI 2016: 19th International Conference, Athens, Greece, October 17-21, 2016, Proceedings, Part III 19},
  pages={397--405},
  year={2016},
  organization={Springer}
}

@article{robben2016,
title = {Simultaneous segmentation and anatomical labeling of the cerebral vasculature},
journal = {Medical Image Analysis},
volume = {32},
pages = {201-215},
year = {2016},
issn = {1361-8415},
doi = {https://doi.org/10.1016/j.media.2016.03.006},
url = {https://www.sciencedirect.com/science/article/pii/S1361841516300056},
author = {David Robben and Engin Türetken and Stefan Sunaert and Vincent Thijs and Guy Wilms and Pascal Fua and Frederik Maes and Paul Suetens}
}

@ARTICLE{tueretken2016,
  author={Türetken, Engin and Benmansour, Fethallah and Andres, Bjoern and Głowacki, Przemysław and Pfister, Hanspeter and Fua, Pascal},
  journal={IEEE Transactions on Pattern Analysis and Machine Intelligence}, 
  title={Reconstructing Curvilinear Networks Using Path Classifiers and Integer Programming}, 
  year={2016},
  volume={38},
  number={12},
  pages={2515-2530},
  doi={10.1109/TPAMI.2016.2519025}
}

@InProceedings{robben2014,
author="Robben, David
and T{\"u}retken, Engin
and Sunaert, Stefan
and Thijs, Vincent
and Wilms, Guy
and Fua, Pascal
and Maes, Frederik
and Suetens, Paul",
editor="Golland, Polina
and Hata, Nobuhiko
and Barillot, Christian
and Hornegger, Joachim
and Howe, Robert",
title="Simultaneous Segmentation and Anatomical Labeling of the Cerebral Vasculature",
booktitle="Medical Image Computing and Computer-Assisted Intervention -- MICCAI 2014",
year="2014",
publisher="Springer International Publishing",
address="Cham",
pages="307--314",
isbn="978-3-319-10404-1"
}

@Article{tueretken2011,
author={T{\"u}retken, Engin
and Gonz{\'a}lez, Germ{\'a}n
and Blum, Christian
and Fua, Pascal},
title={Automated Reconstruction of Dendritic and Axonal Trees by Global Optimization with Geometric Priors},
journal={Neuroinformatics},
year={2011},
month={Sep},
day={01},
volume={9},
number={2},
pages={279-302},
issn={1559-0089},
doi={10.1007/s12021-011-9122-1},
url={https://doi.org/10.1007/s12021-011-9122-1}
}

@InProceedings{tueretken2010,
author="T{\"u}retken, Engin
and Blum, Christian
and Gonz{\'a}lez, Germ{\'a}n
and Fua, Pascal",
editor="Jiang, Tianzi
and Navab, Nassir
and Pluim, Josien P. W.
and Viergever, Max A.",
title="Reconstructing Geometrically Consistent Tree Structures from Noisy Images",
booktitle="Medical Image Computing and Computer-Assisted Intervention -- MICCAI 2010",
year="2010",
publisher="Springer Berlin Heidelberg",
address="Berlin, Heidelberg",
pages="291--299",
isbn="978-3-642-15705-9"
}

@Article{todorov2020,
author={Todorov, Mihail Ivilinov
and Paetzold, Johannes Christian
and Schoppe, Oliver
and Tetteh, Giles
and Shit, Suprosanna
and Efremov, Velizar
and Todorov-V{\"o}lgyi, Katalin
and D{\"u}ring, Marco
and Dichgans, Martin
and Piraud, Marie
and Menze, Bjoern
and Ert{\"u}rk, Ali},
title={Machine learning analysis of whole mouse brain vasculature},
journal={Nature Methods},
year={2020},
month={Apr},
day={01},
volume={17},
number={4},
pages={442-449},
issn={1548-7105},
doi={10.1038/s41592-020-0792-1},
url={https://doi.org/10.1038/s41592-020-0792-1}
}

@article {catmaid2016,
article_type = {journal},
title = {Quantitative neuroanatomy for connectomics in \textit{Drosophila}},
author = {Schneider-Mizell, Casey M and Gerhard, Stephan and Longair, Mark and Kazimiers, Tom and Li, Feng and Zwart, Maarten F and Champion, Andrew and Midgley, Frank M and Fetter, Richard D and Saalfeld, Stephan and Cardona, Albert},
editor = {Calabrese, Ronald L},
volume = 5,
year = 2016,
month = {mar},
pub_date = {2016-03-18},
pages = {e12059},
citation = {eLife 2016;5:e12059},
doi = {10.7554/eLife.12059},
url = {https://doi.org/10.7554/eLife.12059},
journal = {eLife},
issn = {2050-084X},
publisher = {eLife Sciences Publications, Ltd},
}

@article{catmaid2009,
    author = {Saalfeld, Stephan and Cardona, Albert and Hartenstein, Volker and Tomančák, Pavel},
    title = {CATMAID: collaborative annotation toolkit for massive amounts of image data},
    journal = {Bioinformatics},
    volume = {25},
    number = {15},
    pages = {1984-1986},
    year = {2009},
    month = {04},
    issn = {1367-4803},
    doi = {10.1093/bioinformatics/btp266},
    url = {https://doi.org/10.1093/bioinformatics/btp266},
    eprint = {https://academic.oup.com/bioinformatics/article-pdf/25/15/1984/48994215/bioinformatics\_25\_15\_1984.pdf},
}

@Article{Drees2021voreenSkel,
author={Drees, Dominik
and Scherzinger, Aaron
and H{\"a}gerling, Ren{\'e}
and Kiefer, Friedemann
and Jiang, Xiaoyi},
title={Scalable robust graph and feature extraction for arbitrary vessel networks in large volumetric datasets},
journal={BMC Bioinformatics},
year={2021},
month={Jun},
day={26},
volume={22},
number={1},
pages={346},
issn={1471-2105},
doi={10.1186/s12859-021-04262-w},
url={https://doi.org/10.1186/s12859-021-04262-w}
}

@ARTICLE{Drees2019gerome,
  author={Drees, Dominik and Scherzinger, Aaron and Jiang, Xiaoyi},
  journal={IEEE Access}, 
  title={GERoMe-a Method for Evaluating Stability of Graph Extraction Algorithms Without Ground Truth}, 
  year={2019},
  volume={7},
  number={},
  pages={21744-21755},
  doi={10.1109/ACCESS.2019.2898754}
}

@Article{Pampols-Perez2025piezo2,
author={Pampols-Perez, Mireia
and F{\"u}rst, Carina
and S{\'a}nchez-Carranza, Oscar
and Cano, Elena
and Garcia-Contreras, Jonathan Alexis
and Mais, Lisa
and Luo, Wenhan
and Raimundo, Sandra
and Lindberg, Eric L.
and Taube, Martin
and Heuser, Arnd
and Sporbert, Anje
and Kainmueller, Dagmar
and Bernabeu, Miguel O.
and H{\"u}bner, Norbert
and Gerhardt, Holger
and Lewin, Gary R.
and Hammes, Annette},
title={Mechanosensitive PIEZO2 channels shape coronary artery development},
journal={Nature Cardiovascular Research},
year={2025},
month={Jul},
day={01},
volume={4},
number={7},
pages={921-937},
issn={2731-0590},
doi={10.1038/s44161-025-00677-3},
url={https://doi.org/10.1038/s44161-025-00677-3}
}

@Article{Obenaus2017,
author={Obenaus, Andre
and Ng, Michelle
and Orantes, Amanda M.
and Kinney-Lang, Eli
and Rashid, Faisal
and Hamer, Mary
and DeFazio, Richard A.
and Tang, Jiping
and Zhang, John H.
and Pearce, William J.},
title={Traumatic brain injury results in acute rarefication of the vascular network},
journal={Scientific Reports},
year={2017},
month={Mar},
day={22},
volume={7},
number={1},
pages={239},
issn={2045-2322},
doi={10.1038/s41598-017-00161-4},
url={https://doi.org/10.1038/s41598-017-00161-4}
}

@article {cheng2024,
article_type = {journal},
title = {Hemodynamics regulate spatiotemporal artery muscularization in the developing circle of Willis},
author = {Cheng, Siyuan and Xia, Ivan Fan and Wanner, Renate and Abello, Javier and Stratman, Amber N and Nicoli, Stefania},
editor = {Childs, Sarah and Stainier, Didier YR},
volume = 13,
year = 2024,
month = {jul},
pub_date = {2024-07-10},
pages = {RP94094},
citation = {eLife 2024;13:RP94094},
doi = {10.7554/eLife.94094},
url = {https://doi.org/10.7554/eLife.94094},
journal = {eLife},
issn = {2050-084X},
publisher = {eLife Sciences Publications, Ltd},
}

@article {soubeyrand2023,
article_type = {journal},
title = {Three-dimensional imaging of vascular development in the mouse epididymis},
author = {Damon-Soubeyrand, Christelle and Bongiovanni, Antonino and Chorfa, Areski and Goubely, Chantal and Pirot, Nelly and Pardanaud, Luc and Piboin-Fragner, Laurence and Vachias, Caroline and Bravard, Stephanie and Guiton, Rachel and Thomas, Jean-Leon and Saez, Fabrice and Kocer, Ayhan and Tardivel, Meryem and Drevet, Joël R and Henry-Berger, Joelle},
editor = {Kumar, T Rajendra and Zaidi, Mone},
volume = 12,
year = 2023,
month = {jun},
pub_date = {2023-06-13},
pages = {e82748},
citation = {eLife 2023;12:e82748},
doi = {10.7554/eLife.82748},
url = {https://doi.org/10.7554/eLife.82748},
journal = {eLife},
issn = {2050-084X},
publisher = {eLife Sciences Publications, Ltd},
}

@Article{Walek2023,
author={Walek, Konrad W.
and Stefan, Sabina
and Lee, Jang-Hoon
and Puttigampala, Pooja
and Kim, Anna H.
and Park, Seong Wook
and Marchand, Paul J.
and Lesage, Frederic
and Liu, Tao
and Huang, Yu-Wen Alvin
and Boas, David A.
and Moore, Christopher
and Lee, Jonghwan},
title={Near-lifespan longitudinal tracking of brain microvascular morphology, topology, and flow in male mice},
journal={Nature Communications},
year={2023},
month={May},
day={24},
volume={14},
number={1},
pages={2982},
issn={2041-1723},
doi={10.1038/s41467-023-38609-z},
url={https://doi.org/10.1038/s41467-023-38609-z}
}

@InProceedings{Wittmann_2025_CVPR,
    author    = {Wittmann, Bastian and Wattenberg, Yannick and Amiranashvili, Tamaz and Shit, Suprosanna and Menze, Bjoern},
    title     = {vesselFM: A Foundation Model for Universal 3D Blood Vessel Segmentation},
    booktitle = {Proceedings of the Computer Vision and Pattern Recognition Conference (CVPR)},
    month     = {June},
    year      = {2025},
    pages     = {20874-20884}
}

@article{Bumgarner2022vesselvio,
title = {Open-source analysis and visualization of segmented vasculature datasets with VesselVio},
journal = {Cell Reports Methods},
volume = {2},
number = {4},
pages = {100189},
year = {2022},
issn = {2667-2375},
doi = {https://doi.org/10.1016/j.crmeth.2022.100189},
url = {https://www.sciencedirect.com/science/article/pii/S2667237522000443},
author = {Jacob R. Bumgarner and Randy J. Nelson}
}

@inproceedings{
zhu2021deformable,
title={Deformable {\{}DETR{\}}: Deformable Transformers for End-to-End Object Detection},
author={Xizhou Zhu and Weijie Su and Lewei Lu and Bin Li and Xiaogang Wang and Jifeng Dai},
booktitle={International Conference on Learning Representations},
year={2021},
url={https://openreview.net/forum?id=gZ9hCDWe6ke}
}

@InProceedings{carion2020eccv,
author="Carion, Nicolas
and Massa, Francisco
and Synnaeve, Gabriel
and Usunier, Nicolas
and Kirillov, Alexander
and Zagoruyko, Sergey",
editor="Vedaldi, Andrea
and Bischof, Horst
and Brox, Thomas
and Frahm, Jan-Michael",
title="End-to-End Object Detection with Transformers",
booktitle="Computer Vision -- ECCV 2020",
year="2020",
publisher="Springer International Publishing",
address="Cham",
pages="213--229",
isbn="978-3-030-58452-8"
}

@Article{Lyu2022reta,
author={Lyu, Xingzheng
and Cheng, Li
and Zhang, Sanyuan},
title={The RETA Benchmark for Retinal Vascular Tree Analysis},
journal={Scientific Data},
year={2022},
month={Jul},
day={11},
volume={9},
number={1},
pages={397},
issn={2052-4463},
doi={10.1038/s41597-022-01507-y},
url={https://doi.org/10.1038/s41597-022-01507-y}
}

@InProceedings{Shit2021cldice,
    author    = {Shit, Suprosanna and Paetzold, Johannes C. and Sekuboyina, Anjany and Ezhov, Ivan and Unger, Alexander and Zhylka, Andrey and Pluim, Josien P. W. and Bauer, Ulrich and Menze, Bjoern H.},
    title     = {clDice - A Novel Topology-Preserving Loss Function for Tubular Structure Segmentation},
    booktitle = {Proceedings of the IEEE/CVF Conference on Computer Vision and Pattern Recognition (CVPR)},
    month     = {June},
    year      = {2021},
    pages     = {16560-16569}
}

@InProceedings{lin2014coco,
author="Lin, Tsung-Yi
and Maire, Michael
and Belongie, Serge
and Hays, James
and Perona, Pietro
and Ramanan, Deva
and Doll{\'a}r, Piotr
and Zitnick, C. Lawrence",
editor="Fleet, David
and Pajdla, Tomas
and Schiele, Bernt
and Tuytelaars, Tinne",
title="Microsoft COCO: Common Objects in Context",
booktitle="Computer Vision -- ECCV 2014",
year="2014",
publisher="Springer International Publishing",
address="Cham",
pages="740--755",
isbn="978-3-319-10602-1"
}

@Article{Foucart2023,
author={Foucart, Adrien
and Debeir, Olivier
and Decaestecker, Christine},
title={Panoptic quality should be avoided as a metric for assessing cell nuclei segmentation and classification in digital pathology},
journal={Scientific Reports},
year={2023},
month={May},
day={27},
volume={13},
number={1},
pages={8614},
issn={2045-2322},
doi={10.1038/s41598-023-35605-7},
url={https://doi.org/10.1038/s41598-023-35605-7}
}

@INPROCEEDINGS{li2023ted,
  author={Li, Wenjian and Wu, Guannan and Li, Huiqi},
  booktitle={2023 IEEE 18th Conference on Industrial Electronics and Applications (ICIEA)}, 
  title={Similarity evaluation of retinal vascular network based on tree edit distance}, 
  year={2023},
  volume={},
  number={},
  pages={722-727},
  doi={10.1109/ICIEA58696.2023.10241828}}

@article{matula2015tra,
    doi = {10.1371/journal.pone.0144959},
    author = {Matula, Pavel AND Maška, Martin AND Sorokin, Dmitry V. AND Matula, Petr AND Ortiz-de-Solórzano, Carlos AND Kozubek, Michal},
    journal = {PLOS ONE},
    publisher = {Public Library of Science},
    title = {Cell Tracking Accuracy Measurement Based on Comparison of Acyclic Oriented Graphs},
    year = {2015},
    month = {12},
    volume = {10},
    url = {https://doi.org/10.1371/journal.pone.0144959},
    pages = {1-19},
    number = {12},

}

@InProceedings{fisbe,
    author    = {Mais, Lisa and Hirsch, Peter and Managan, Claire and Kandarpa, Ramya and Rumberger, Josef Lorenz and Reinke, Annika and Maier-Hein, Lena and Ihrke, Gudrun and Kainmueller, Dagmar},
    title     = {FISBe: A Real-World Benchmark Dataset for Instance Segmentation of Long-Range Thin Filamentous Structures},
    booktitle = {Proceedings of the IEEE/CVF Conference on Computer Vision and Pattern Recognition (CVPR)},
    month     = {June},
    year      = {2024},
    pages     = {22249-22259}
}

@article{schneider2012tissue,
  title={Tissue metabolism driven arterial tree generation},
  author={Schneider, Matthias and Reichold, Johannes and Weber, Bruno and Sz{\'e}kely, G{\'a}bor and Hirsch, Sven},
  journal={Medical image analysis},
  volume={16},
  number={7},
  pages={1397--1414},
  year={2012},
  publisher={Elsevier}
}

@software{Silversmith_Kimimaro_Skeletonize_densely_2021,
author = {Silversmith, William and Bae, J. Alexander and Li, Peter H. and Wilson, A.M.},
doi = {10.5281/zenodo.5539913},
month = sep,
title = {{Kimimaro: Skeletonize densely labeled 3D image segmentations}},
version = {3.0.0},
year = {2021}
}

@article{liu2021assessment,
  title={Assessment of Vascular Network Connectivity of Hepatocellular Carcinoma Using Graph-Based Approach},
  author={Liu, Qiaoyu and Zhang, Boyu and Wang, Luna and Zheng, Rencheng and Qiang, Jinwei and Wang, He and Yan, Fuhua and Li, Ruokun},
  journal={Frontiers in Oncology},
  volume={11},
  pages={668874},
  year={2021}
}

@Article{Malin-Mayor2023,
author={Malin-Mayor, Caroline
and Hirsch, Peter
and Guignard, Leo
and McDole, Katie
and Wan, Yinan
and Lemon, William C.
and Kainmueller, Dagmar
and Keller, Philipp J.
and Preibisch, Stephan
and Funke, Jan},
title={Automated reconstruction of whole-embryo cell lineages by learning from sparse annotations},
journal={Nature Biotechnology},
year={2023},
month={Jan},
day={01},
volume={41},
number={1},
pages={44-49},
issn={1546-1696},
doi={10.1038/s41587-022-01427-7},
url={https://doi.org/10.1038/s41587-022-01427-7}
}

@InProceedings{patchperpix,
author="Mais, Lisa
and Hirsch, Peter
and Kainmueller, Dagmar",
editor="Vedaldi, Andrea
and Bischof, Horst
and Brox, Thomas
and Frahm, Jan-Michael",
title="PatchPerPix for Instance Segmentation",
booktitle="Computer Vision -- ECCV 2020",
year="2020",
publisher="Springer International Publishing",
address="Cham",
pages="288--304",
isbn="978-3-030-58595-2"
}

@inproceedings{
lux2025topograph,
title={Topograph: An Efficient Graph-Based Framework for Strictly Topology Preserving Image Segmentation},
author={Laurin Lux and Alexander H Berger and Alexander Weers and Nico Stucki and Daniel Rueckert and Ulrich Bauer and Johannes C. Paetzold},
booktitle={The Thirteenth International Conference on Learning Representations},
year={2025},
url={https://openreview.net/forum?id=Q0zmmNNePz}
}

@InProceedings{kirchhoff2024,
author="Kirchhoff, Yannick
and Rokuss, Maximilian R.
and Roy, Saikat
and Kovacs, Balint
and Ulrich, Constantin
and Wald, Tassilo
and Zenk, Maximilian
and Vollmuth, Philipp
and Kleesiek, Jens
and Isensee, Fabian
and Maier-Hein, Klaus",
editor="Leonardis, Ale{\v{s}}
and Ricci, Elisa
and Roth, Stefan
and Russakovsky, Olga
and Sattler, Torsten
and Varol, G{\"u}l",
title="Skeleton Recall Loss for Connectivity Conserving and Resource Efficient Segmentation of Thin Tubular Structures",
booktitle="Computer Vision -- ECCV 2024",
year="2024",
publisher="Springer Nature Switzerland",
address="Cham",
pages="218--234",
isbn="978-3-031-72980-5"
}

@Article{RiosCoronado2025,
author={Rios Coronado, Pamela E.
and Zhou, Jiayan
and Fan, Xiaochen
and Zanetti, Daniela
and Naftaly, Jeffrey A.
and Prabala, Pratima
and Mart{\'i}nez Jaimes, Azalia M.
and Farah, Elie N.
and Kundu, Soumya
and Deshpande, Salil S.
and Evergreen, Ivy
and Kho, Pik Fang
and Ma, Qixuan
and Hilliard, Austin T.
and Abramowitz, Sarah
and Pyarajan, Saiju
and Dochtermann, Daniel
and Damrauer, Scott M.
and Chang, Kyong-Mi
and Levin, Michael G.
and Winn, Virginia D.
and Pa{\c{s}}ca, Anca M.
and Plomondon, Mary E.
and Waldo, Stephen W.
and Tsao, Philip S.
and Kundaje, Anshul
and Chi, Neil C.
and Clarke, Shoa L.
and Red-Horse, Kristy
and Assimes, Themistocles L.},
title={<em>CXCL12</em> drives natural variation in coronary artery anatomy across diverse populations},
journal={Cell},
year={2025},
month={Apr},
day={03},
publisher={Elsevier},
volume={188},
number={7},
pages={1784-1806.e22},
issn={0092-8674},
doi={10.1016/j.cell.2025.02.005},
url={https://doi.org/10.1016/j.cell.2025.02.005}
}

@Article{Sheridan2023lsd,
author={Sheridan, Arlo
and Nguyen, Tri M.
and Deb, Diptodip
and Lee, Wei-Chung Allen
and Saalfeld, Stephan
and Turaga, Srinivas C.
and Manor, Uri
and Funke, Jan},
title={Local shape descriptors for neuron segmentation},
journal={Nature Methods},
year={2023},
month={Feb},
day={01},
volume={20},
number={2},
pages={295-303},
issn={1548-7105},
doi={10.1038/s41592-022-01711-z},
url={https://doi.org/10.1038/s41592-022-01711-z}
}

@article{sexton2025svt,
author = {Zachary A. Sexton  and Dominic Rütsche  and Jessica E. Herrmann  and Andrew R. Hudson  and Soham Sinha  and Jianyi Du  and Daniel J. Shiwarski  and Anastasiia Masaltseva  and Fredrik Samdal Solberg  and Jonathan Pham  and Jason M. Szafron  and Sean M. Wu  and Adam W. Feinberg  and Mark A. Skylar-Scott  and Alison L. Marsden },
title = {Rapid model-guided design of organ-scale synthetic vasculature for biomanufacturing},
journal = {Science},
volume = {388},
number = {6752},
pages = {1198-1204},
year = {2025},
doi = {10.1126/science.adj6152},
URL = {https://www.science.org/doi/abs/10.1126/science.adj6152},
eprint = {https://www.science.org/doi/pdf/10.1126/science.adj6152}}

@misc{luo2024parse,
      title={Efficient automatic segmentation for multi-level pulmonary arteries: The PARSE challenge}, 
      author={Gongning Luo and Kuanquan Wang and Jun Liu and Shuo Li and Xinjie Liang and Xiangyu Li and Shaowei Gan and Wei Wang and Suyu Dong and Wenyi Wang and Pengxin Yu and Enyou Liu and Hongrong Wei and Na Wang and Jia Guo and Huiqi Li and Zhao Zhang and Ziwei Zhao and Na Gao and Nan An and Ashkan Pakzad and Bojidar Rangelov and Jiaqi Dou and Song Tian and Zeyu Liu and Yi Wang and Ampatishan Sivalingam and Kumaradevan Punithakumar and Zhaowen Qiu and Xin Gao},
      year={2024},
      eprint={2304.03708},
      archivePrefix={arXiv},
      primaryClass={eess.IV},
      url={https://arxiv.org/abs/2304.03708}, 
}

@book{diestel2017graph,
  author    = {Reinhard Diestel},
  title     = {Graph Theory},
  edition   = {5},
  publisher = {Springer},
  series    = {Graduate Texts in Mathematics},
  year      = {2017},
  isbn      = {978-3-319-41320-4}
}

@Article{Breiman2001RF,
author={Breiman, Leo},
title={Random Forests},
journal={Machine Learning},
year={2001},
month={Oct},
day={01},
volume={45},
number={1},
pages={5-32},
issn={1573-0565},
doi={10.1023/A:1010933404324},
url={https://doi.org/10.1023/A:1010933404324}
}

@article{Frangi2000,
author = {Frangi, Alejandro and Niessen, W.J. and Vincken, Koen and Viergever, Max},
year = {2000},
month = {02},
pages = {},
title = {Multiscale Vessel Enhancement Filtering},
volume = {1496},
journal = {Med. Image Comput. Comput. Assist. Interv.}
}

@inproceedings{hirsch2020_an_auxil_task_for_learn_nucle_segme_in_3d,
  author =       {Hirsch, Peter and Kainmueller, Dagmar},
  title =        {An Auxiliary Task for Learning Nuclei Segmentation in 3D
                  Microscopy Images},
  booktitle =    {International Conference on Medical Imaging with Deep
                  Learning, {MIDL} 2020, 6-8 July 2020, Montr{\'{e}}al, QC,
                  Canada},
  series =       {Proceedings of Machine Learning Research},
  volume =       121,
  pages =        {304--321},
  publisher =    {{PMLR}},
  year =         2020,
  url =          {http://proceedings.mlr.press/v121/hirsch20a.html},
  bibsource =    {dblp computer science bibliography, https://dblp.org}
}

@article{isensee2021nnu,
  title={nnU-Net: a self-configuring method for deep learning-based biomedical image segmentation},
  author={Isensee, Fabian and Jaeger, Paul F and Kohl, Simon AA and Petersen, Jens and Maier-Hein, Klaus H},
  journal={Nature methods},
  volume={18},
  number={2},
  pages={203--211},
  year={2021},
  publisher={Nature Publishing Group US New York}
}

\appendix
\section{Extended Methods}
\begin{figure}
  \centering
  \subfigure[\,]{%
    \includegraphics[height=3cm]{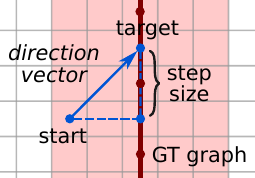}
    \label{fig:direction_vectors_def}
  }
  \quad
  \subfigure[\,]{%
    \includegraphics[height=3cm]{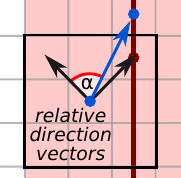}
    \label{fig:angular_difference}
  }
  \caption{\textbf{Direction vectors generation and angular difference penalty used in modified TEASAR.}
  (a) Direction vectors (blue) are generated by first identifying the closest point on the ground-truth graph (dark red) and then stepping a fixed distance toward the root along the graph. Since the step size is constant across all voxel locations, direction vectors near the centerline exhibit smaller magnitudes, while those near the vessel boundary have larger magnitudes.
  (b) A penalty is assigned based on the angular difference between the predicted direction vector at the current voxel and the relative direction vector to each of its neighboring voxels (i.e., the ``walk direction''). Lower penalties correspond to stronger alignment between the predicted and actual tracing directions, encouraging paths that follow the learned vessel orientation.}
  \label{fig:direction_vectors}
\end{figure}

\begin{algorithm}[t]
\caption{Direction Vector Generation}
\label{alg:vector_generation}

\KwIn{Vessel segmentation mask $M$ and the corresponding skeleton graph $G=(V,E)$ with each edge $e\in E$ having a radius $r_e > 0$ assigned}

\textbf{Parameters:} \texttt{step\_size}$>0$

\KwOut{$\texttt{vector\_field}$ of direction vectors}

Fit $G$ to $M$ by rescaling

Find leaf nodes in $G$

Select the root of each connected component to be the leaf node with maximal radius

Direct the edges in each connected component tree such that they are directed towards the root.

\ForEach{\texttt{foreground\_voxel} of $M$}{
    Find \texttt{nearest\_edge} $\in E$ to \texttt{foreground\_voxel} in $G$
    
    Find \texttt{closest\_point} to \texttt{foreground\_voxel} on \texttt{nearest\_edge} 
    
    $d\gets$ distance between \texttt{foreground\_voxel} and \texttt{closest\_point}
    
    \If{$d \leq r_{\text{\texttt{nearest\_edge}}}$}{
        \texttt{target\_point} $\gets$ move \texttt{closest\_point} for \texttt{step\_size} along $G$ towards the root
        
        \texttt{direction\_vector} $\gets$ \texttt{target\_point} - \texttt{foreground\_voxel}
        
        Store \texttt{direction\_vector} in \texttt{vector\_field}
    }
}
\Return{$\texttt{vector\_field}$}\;

\end{algorithm}

\subsection{Automated Root Detection}\label{sec:automated_root_detection}

Vesselpose extracts a centerline graph using an adaptation of the TEASAR algorithm, which requires an initial set of vessel roots. In practice, these roots are annotated manually, a procedure that is both slow and labor-intensive. An automated strategy would remove this bottleneck, but detecting roots directly from the binary foreground mask is unreliable.

The voxel-wise direction field predicted by Vesselpose provides a direct workaround: roots correspond to sinks of this 3D flow field. A virtual particle is hereby initialized at every foreground voxel \(p\) whose distance to the background is at least \(r_{\min}\). The particle position is set to \(x_0 := p\) and updated iteratively by $x_i := x_{i-1} + \lambda\, v_{i-1}$, 
where \(v_{i-1}\) is the interpolated direction vector at \(x_{i-1}\) and \(\lambda>0\) is a step-size parameter. A position \(x_i\) is considered a sink when the displacement magnitude \(\lambda \lVert v_i \rVert\) falls below a tolerance \(\tau>0\). After \(N\) iterations, all detected sinks are collected and used as candidate vessel roots for initializing the adapted TEASAR procedure.

To evaluate the effectiveness of automated root detection we apply this method to the three validation data of the Multi-tree synthetic dataset and report how many manually annotated ground-truth roots are correctly detected. A root is considered to be correctly detected if it lies within a distance of $2$ voxels from a calculated sink. For our experiments we further chose the following parameters: number of steps $N=50$, step size $\lambda=1.0$, tolerance $\tau= 0.1$ and min-radius threshold $r_{\text{min}}=3.0$.

As a result, the automated root detection algorithm is able to detect \emph{all} of the ground-truth roots present in all three volume samples, while producing additional false positive roots.
However, the additional false positive roots do not pose a problem to the TEASAR-based centerline generation since these can be filtered out during multi-root processing as described in \ref{sec:teasar}.

\subsection{Adaptive masking}\label{sec:adaptive masking}
Standard TEASAR excludes processed areas using simple linear thresholding with a fixed scale and constant. However, in vesselpose these parameters vary with respect to the local radius as described in \ref{sec:teasar}. 
Given predefined parameter ranges for $scale \in [s_{\min}, s_{\max}]$ and $const \in [c_{\min}, c_{\max}]$, and radius bounds $[r_{\min}, r_{\max}]$, we compute the normalized radius fraction
\begin{equation}
\alpha_r =
\frac{\operatorname{r} - r_{\min}}
{r_{\max} - r_{\min}}
\end{equation}
The scale and constant are then interpolated:
\begin{equation}
\text{scale}(r) = s_{\min} + \alpha_r(s_{\max} - s_{\min}) ~,~
\text{const}(r) = c_{\min} + \alpha_r(c_{\max} - c_{\min})
\end{equation}
The adaptive masking distance is finally defined as
\begin{equation}
d(r) = \text{scale}(r)\cdot r + \text{const}(r)
\end{equation}

\section{Extended Evaluation}

\subsection{Hierarchical Matching}
In~\algorithmref{alg:hierarchical-matching} we demonstrate the steps involved in our hierarchical matching which takes the node label and its semantic into consideration rather than just the spatial proximity.

\begin{algorithm}[t]
\caption{Hierarchical matching between $G$ and $P$}
\label{alg:hierarchical-matching}
\DontPrintSemicolon

\KwIn{Directed acyclic graphs $G$ and $P$;
Each node in each graph labelled as belonging to semantic class "root", "leaf", "branching point" or "intermediate point"; Each node in each graph has an associated 3D position; maximum distance for two points being matched $d_{max}$}
\KwOut{\texttt{match\_dict}}

Initialize empty \texttt{match\_dict}

\ForEach {node $v$ in $G$} {Determine the set $W_v$ of closest nodes in $P$ that lie within distance $d_{max}$ to $v$\;

Determine subset $W_{v,sem} \subseteq W_v$ of nodes with same semantic class as $v$\;

Sort each subset by distance to $v$

Append to form sorted list $W_v^{sorted} = [W_{v,sem},\ W_v \backslash W_{v,sem}]$
}

Sort root nodes $r$ of $G$ analogously, i.e.: Primary sorting criterion: non-empty $W_{r,sem}$ before all others; secondary  criterion: distance of closest $w \in W_r$

\ForEach{$r$ in sorted list of root nodes in $G$}{
    \ForEach{class label $c$ in sorted list $["root", "branching\ point" or "leaf", "intermediate\ point"]$}{
    \ForEach{node $v$ with class label $c$ in depth-first traversal of $G$ from $r$}{
        \If{class label of $v$ is $"root"$}{
            remove all elements $w$ in $W_v^{sorted}$ if parent of $w$ is already matched to another tree than $r$\;
        }
        \If{$W_v^{sorted}$ is not empty} {
            \If{class label of $v$ is no $"root"$} {
                get mask $m_v$ with true elements if parent of $w \in W_v^{sorted}$ is matched to same tree as $r$\;
                
                \If{$W_v^{sorted}[m_v]$ is not empty}{
                    update $W_v^{sorted}$ with $W_v^{sorted}[m_v]$\;
                }
            }

            pick first $w$ in $W_v^{sorted}$\;
            
            store $(v,w)$ in \texttt{match\_dict}\;
            
            \ForEach{not yet visited node $v'$ in $G$}{remove $w$ from $W_{v'}^{sorted}$}
        }
    }
}}
\Return{\texttt{match\_dict}}\;
\end{algorithm}

\subsection{Related work: Commonly used metrics}\label{sec:metrics_discussion}
A unified approach for comparing graph structures is still lacking \citet{Lyu2022reta}, and numerous evaluation strategies have emerged across the biomedical literature.
On one hand, there are \textit{detection- and segmentation-based metrics} for comparing centerlines, branching points, or graph edges, which we find unsuitable for graph-based evaluation:
For example, clDice~\cite{Shit2021cldice} does not reflect topology, as it operates at the pixel level and fails to penalize structural changes like loops or disconnections---errors that may drastically alter topology but only minimally impact clDice scores as few pixels are missing or added.
The same limitation applies to precision, recall, and F1 scores when applied on pixel level.
If mean average precision (mAP)\cite{lin2014coco} is used, as in \citet{vesselformer2024,trexplorer2024}, nodes and edges are compared based on their overlap of the bounding boxes.
However, as shown in \citet{Foucart2023}, intersection-over-union (IoU) is not well-suited for small objects, especially in 3D, where meeting high IoU thresholds becomes impractical.

On the other hand, \textit{graph similarity measures} are commonly used in related studies \cite{vesselformer2024,trexplorer2024,Drees2019gerome,Drees2021voreenSkel}.
However, the Street Mover's Distance (SMD) does not preserve connectivity information as it converts graph to point clouds for distance computation.
Additionally, SMD is sensitive to resampling and hyperparameters, making it less reliable for topological evaluation.
In contrast, Betti numbers provide a topological perspective, but, e.g., Betti-0 is too coarse, as it does not capture topological errors within individual trees \cite{lux2025topograph}.
Instead, precision, recall, and F1 scores---when applied at the graph level, particularly for edges, as in \citet{Drees2019gerome,Drees2021voreenSkel}---provide meaningful insights into topological correctness.
However, these metrics do not account for the structural impact of individual errors: missing or spurious edges may vary greatly in severity depending on how drastically they alter the topology of the graph, a nuance not captured by the F1 score. 

Another family of metrics includes \textit{tree edit distance (TED)-based measures} like \citet{li2023ted,matula2015tra}, which quantify the number of operations (e.g., node or edge insertions/deletions) needed to transform the predicted graph into the ground-truth.
These metrics offer an intuitive and meaningful notion of similarity, but we have not seen them commonly used in related work.
For example, \citet{li2023ted} introduces a TED variant specifically for vasculature graphs, but it penalizes heavily false merges and splits between two trees as it penalizes the falsely merged tree twice by adding the counts for removing every falsely added node and edge; and then for creating the missing tree from scratch.
Another intuitive measure is the TRA metric \cite{matula2015tra}, originally designed for evaluating cell lineages, but it is not directly applicable here as it relies on IoU-based matching.

Our observations are consistent with \citet{Lyu2022reta}, who conclude that selecting a reliable and unbiased evaluation metric remains an open problem in the community. With this work, we aim to contribute to this discussion by proposing a greedy hierarchical matching strategy and introducing false splits and false merges as topology-aware measures for robust assessment of vessel graph reconstructions. Nevertheless, we believe that a more systematic analysis of how node matching and sampling choices influence evaluation outcomes is still needed to establish common recommendations—an important direction for future work, but beyond the scope of this paper.

\begin{table}
    \centering
    \caption{\label{tab:our_results}Quantitative comparison of our hierarchical matching with greedy one-to-one matching and optimal Hungarian matching.
    Lower false merge (FM) and false split (FS) values indicate better topology preservation during matching.
    Results are shown for Vesselpose on the validation set of the Multi-Tree Synthetic dataset.
    For this analysis, graphs were resampled to include only roots, branching points, and end nodes.}
    \begin{tblr}{width=\linewidth,rows={abovesep=1pt,belowsep=1pt}}
        \midrule
        \SetCell[r=2]{l}{Matching} & \SetCell[c=3]{c}{Edges} & & & \SetCell[c=2]{c}{FM} & & \SetCell[c=2]{c}{FS} & \\
        \cmidrule[lr]{2-4} \cmidrule[lr]{5-6} \cmidrule[lr]{7-8}
        & F1$\uparrow$ & Prec$\uparrow$ & Rec$\uparrow$ & Rel.$\downarrow$ & Abs.$\downarrow$ & Rel.$\downarrow$ & Abs.$\downarrow$\\
        \midrule
        Greedy & 0.80 & 0.81 & 0.79 & 0.01 & 49 & 0.01 & 48.6 \\
        Hungarian & \textbf{0.81} & \textbf{0.82} & \textbf{0.80} & 0.02 & 62.6 & 0.01 & 62.3\\
        Hierarchical (Ours) & 0.80 & 0.81 & 0.79 & \textbf{0.007} & \textbf{29.7}& \textbf{0.007} & \textbf{29.3}\\
        \midrule
        \label{tab:matching_comparison}
    \end{tblr}
\end{table}

\section{Extended Experiments}\label{sec:ext_experiments}
We conducted all experiments on an HPC cluster using a single NVIDIA H100 GPU. Each model is trained for approximately 3 days. We allocate 200 GB RAM per job, although CPU and memory demands are modest since training relied mainly on GPU computation with Zarr-based random crop loading. The models are trained using PyTorch on a CUDA-enabled Linux environment.

\subsection{Single-Tree Synthetic Data}\label{sec:ext_single_tree_synthetic}
The dataset comprises 500 volumes of size $256^3$ voxels and is split into training, validation, and test sets.
The training, validation and test split contain 368, 32 and 100 samples respectively. The network is trained on randomly sampled input patches of size $128^3$ voxels with intensity shift augmentation.

\begin{figure}[t]
  \centering
  \subfigure[Raw]{%
    \includegraphics[
      width=0.31\linewidth
    ]{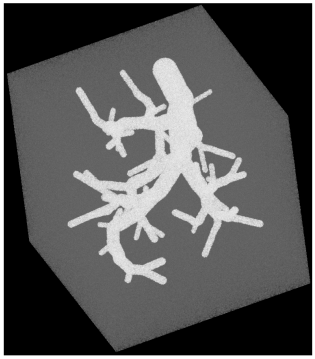}
  }
  \hfill
  \subfigure[Ground-truth]{%
    \includegraphics[
      width=0.31\linewidth]{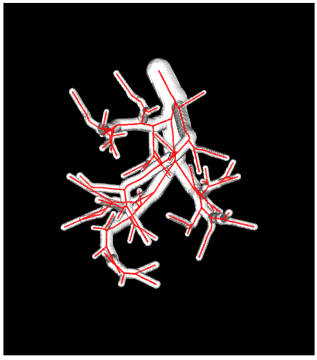}
  }
  \hfill
  \subfigure[Ours]{%
    \includegraphics[
      width=0.305\linewidth]{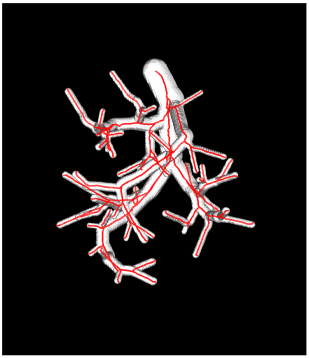}
  }
  \caption{\textbf{Qualitative comparison for single-tree synthetic.}
  A 3D rendering of one of the samples. (a) shows the raw image (b) Ground-truth skeletons overlaid on the segmentation mask (c) our predicted skeleton overlaid on the predicted binary segmentation.
  Our method (c) produces a reconstruction that closely matches the ground-truth (b), capturing fine structures and maintaining topological consistency.
  }
  \label{fig:qualitative_syn_st}
\end{figure}

\subsection{Parse 2022 Challenge}\label{sec:ext_parse}
The data has a in-plane size of $512\times 512$ pixels and its z-stack comprises between 295 and 390 slices.
The training, validation and test split contain 72, 8 and 20 samples respectively.
Following \citet{trexplorer_super2025}, all volumes are resampled to an isotropic resolution of $0.5mm$.
For both training and inference, we use input sizes of $256^3$ voxels and do not apply any data augmentation. 

\begin{figure}[t]
  \centering
  \subfigure[Raw]{%
    \includegraphics[
      trim=140pt 0pt 200pt 100pt,
      clip,
      width=0.31\linewidth
    ]{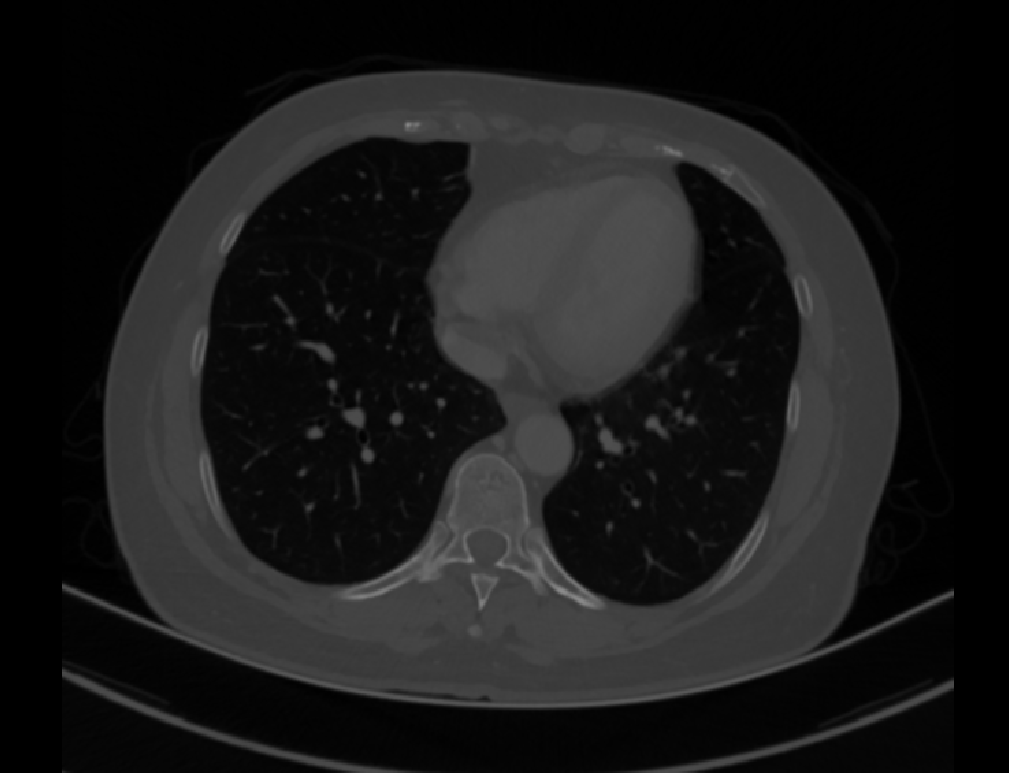}
  }
  \hfill
  \subfigure[Ground-truth]{%
    \includegraphics[trim=140pt 0pt 200pt 100pt,
      clip,
      width=0.31\linewidth]{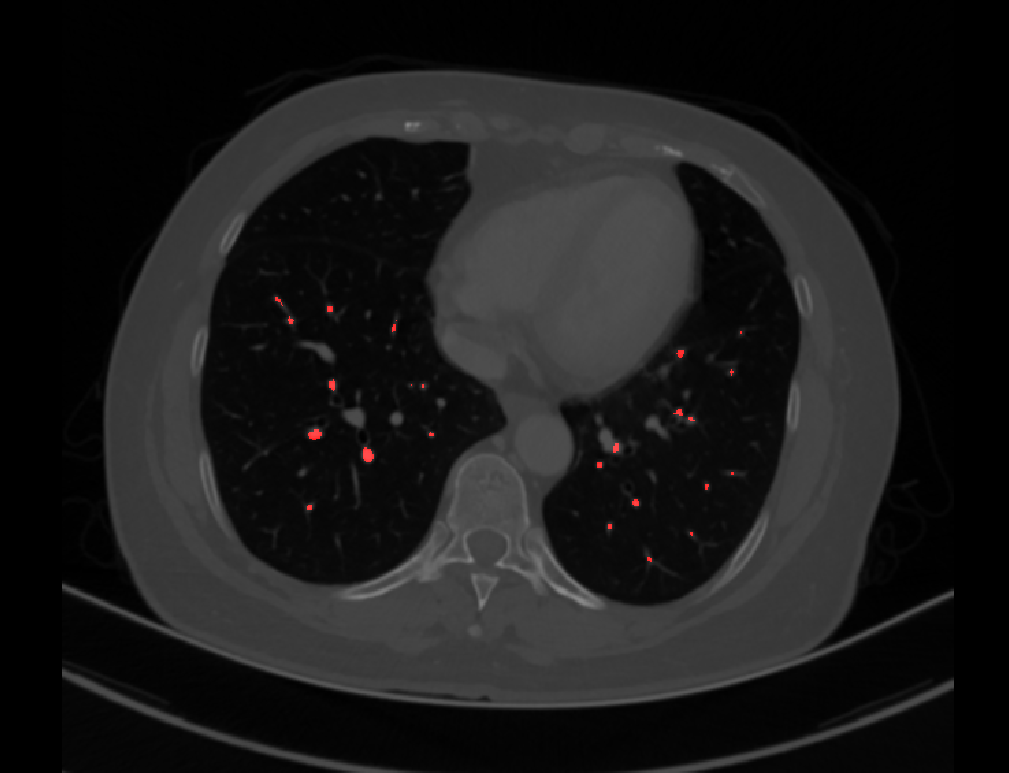}
  }
  \hfill
  \subfigure[Ours]{%
    \includegraphics[trim=140pt 0pt 200pt 100pt,
      clip,
      width=0.31\linewidth]{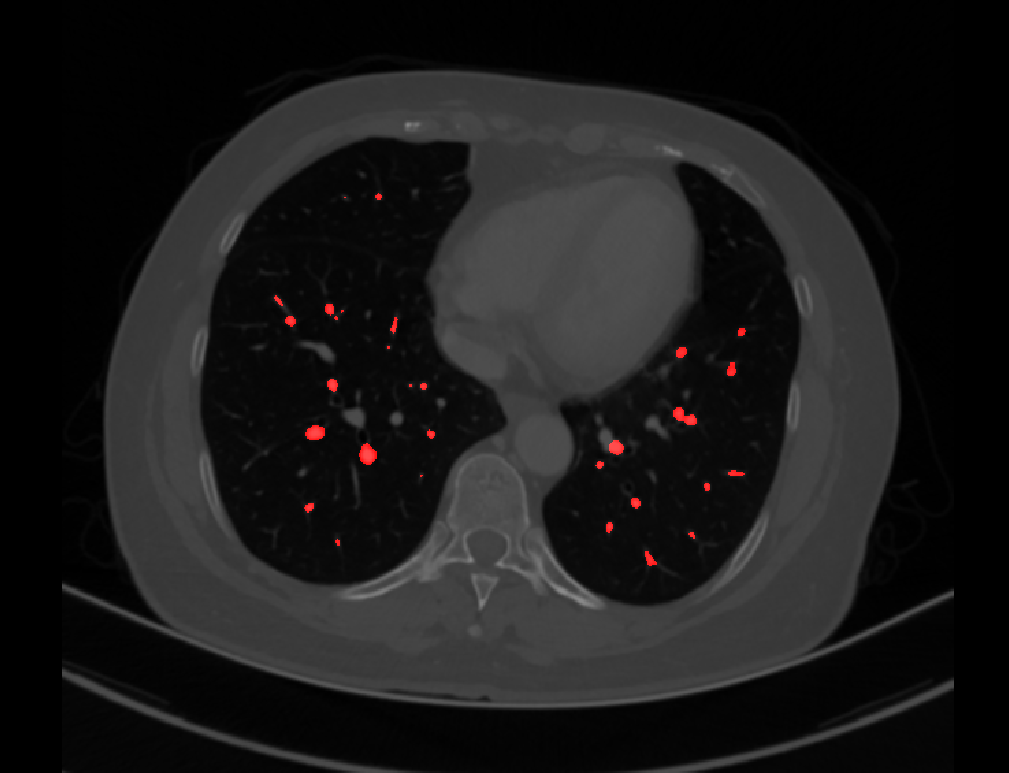}
  }\\
  \subfigure[Raw]{%
    \includegraphics[width=0.31\linewidth]{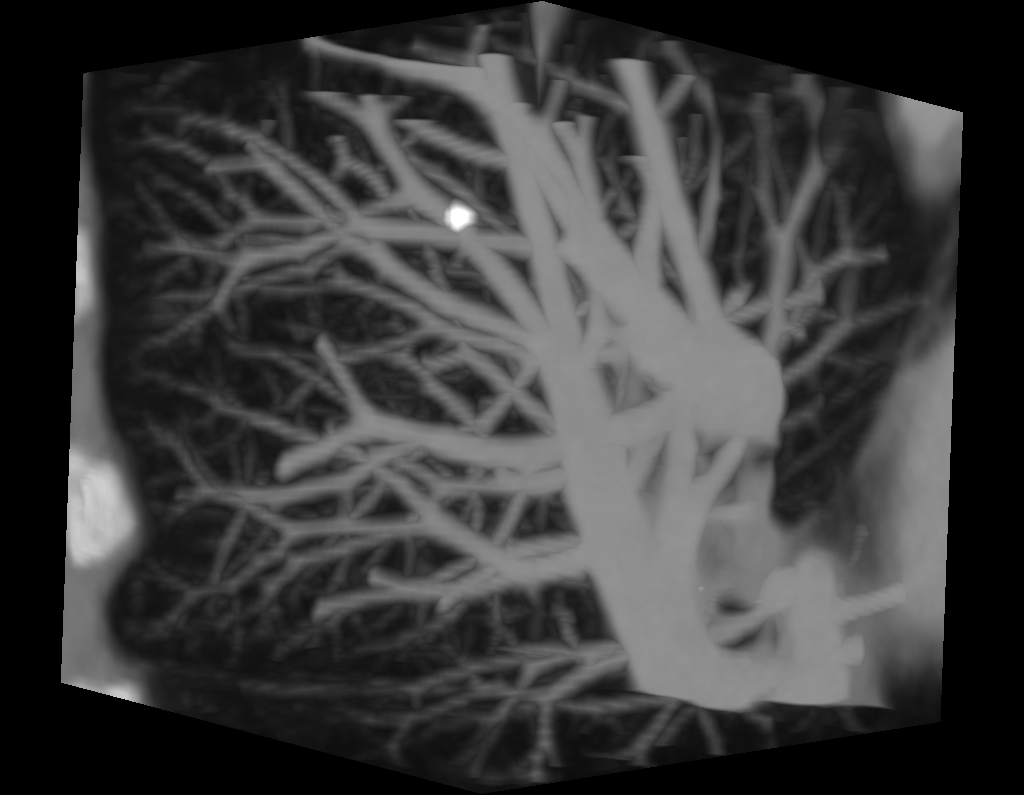}
  }
  \hfill
  \subfigure[Ground-truth]{%
    \includegraphics[width=0.31\linewidth]{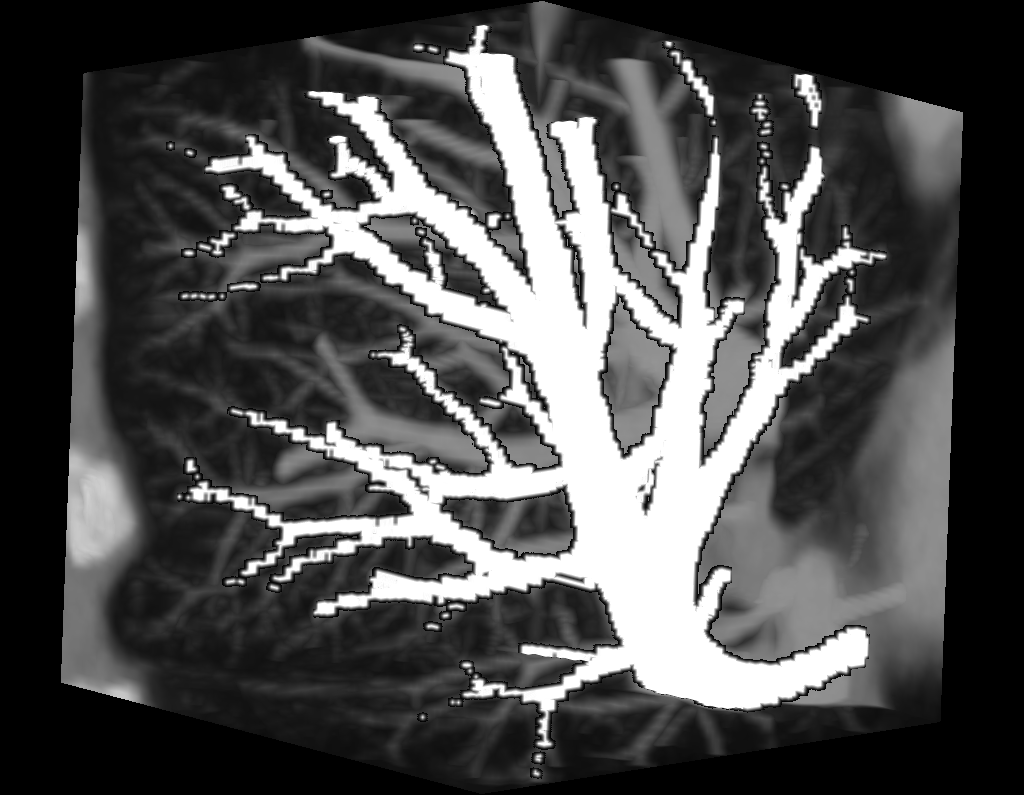}
  }
  \hfill
  \subfigure[Ours]{%
    \includegraphics[width=0.295\linewidth]{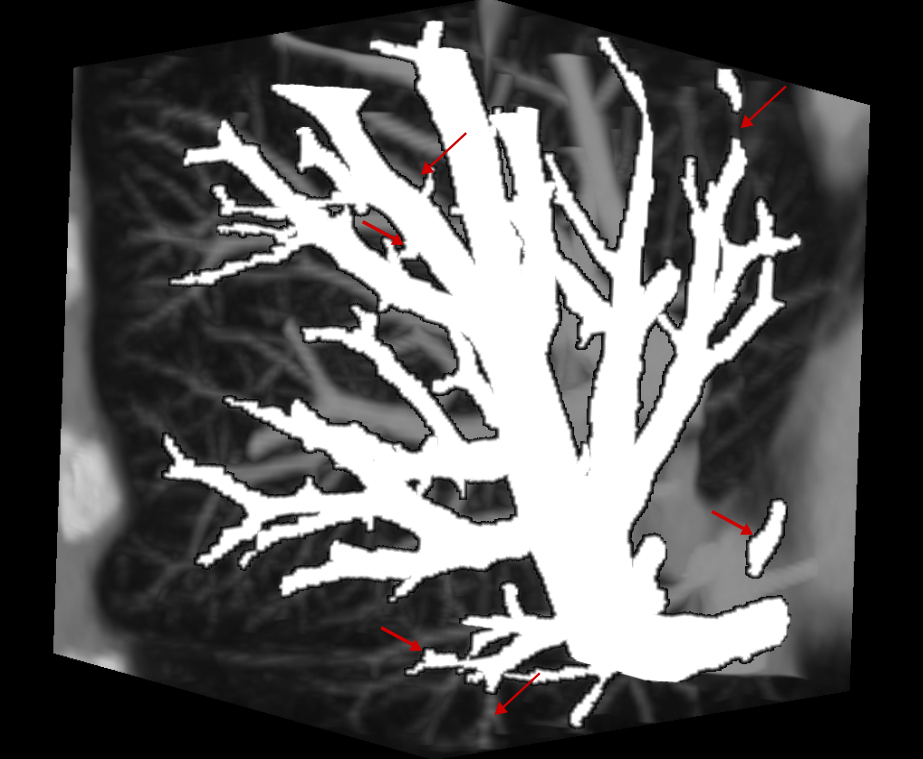}
  }
  \caption{\textbf{Qualitative comparison of PARSE2022 segmentation.}
  First row: A 2D slice from one sample. (a) shows the raw image, while (b) and (c) show the raw image overlaid with the provided ground-truth segmentation and our segmentation, respectively.
  Second row: A 3D crop from the same sample.
  (d) shows the raw image, and (e) and (f) show the raw image overlaid with the ground-truth and our segmentation, respectively.
  Both segmentation masks miss vessel segments that are visible in the raw data and contain disconnected components. Some failure cases in our segmentation are highlighted in red arrows(f).
  }
  \label{fig:qualitative_parse}
\end{figure}

\subsection{Multi-tree Synthetic Data}
This dataset is obtained from \cite{tetteh2019_deepvesselnet}.
Each volume measures $325 \times 304 \times 600$ voxels.
We follow the information provided in \citet{trexplorer2024,vesselformer2024} about how many samples were used in which data split and utilized the first 50 volumes. The training, validation and test split contain 37, 3 and 10 samples respectively.
Training and inference are performed with a input size of $256^3$ voxels, using intensity shifts and masked out crops as data augmentations. For this analysis, graphs are resampled to include only roots, branching points, and end nodes.

\subsection{Micro-CT Heart data}\label{sec:ext_micro_ct}
This dataset contains samples from both preeclamptic and healthy animals.
The individual samples measure approximately $1300 \times 1300 \times 1700$ voxels at a resolution of $12\mu m$.
To obtain foreground masks for fine-tuning the U-Net on the micro-CT heart data, we train a random forest classifier \cite{Breiman2001RF} with 100 trees and a maximum depth of 10.
We include Frangi vesselness features \cite{Frangi2000} as additional input features for the random forest classifier and manually annotate a subset of vessels as ground-truth.
We then use the resulting foreground masks to fine-tune the U-Net model which was pre-trained with the synthetic multi-tree dataset.
We fine-tune the U-Net by freezing all but the final layer to predict the foreground mask.

A key challenge associated with this dataset is its large size and the presence of substantial heart chambers, which occupy much of the volume.
Thus, we downsample the data by a factor of 0.5 in each dimension and apply a heuristic approach to remove the chambers by eliminating the largest connected component in each 2D segmentation slice. 
Fine-tuning and inference are performed with an input size of $256^3$ voxels.

\begin{figure}[t]
  \centering
  \subfigure[Ground-Truth]{%
    \includegraphics[
      width=0.32\linewidth
    ]{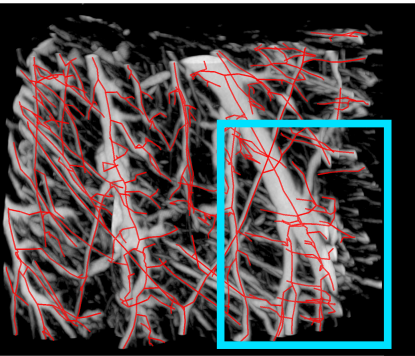}
  }
  \hfill
  \subfigure[Vesselpose]{%
    \includegraphics[
      width=0.31\linewidth]{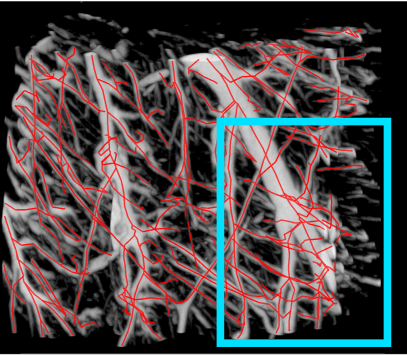}
  }
  \hfill
  \subfigure[U-Net+TEASAR]{%
    \includegraphics[
      width=0.32\linewidth]{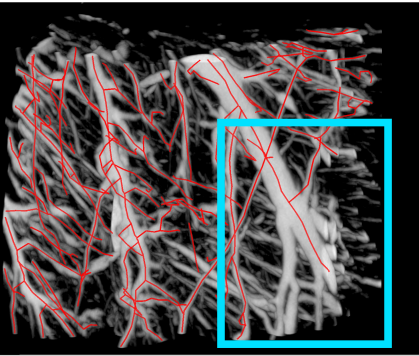}
  }
  \caption{\textbf{Qualitative results of the micro-CT data.}
  Illustrated is one 3D annotated crop from the raw micro-CT test data together with varying vessel skeleton graphs in red:
(a) shows the annotated ground-truth skeleton; (b) shows the results of our proposed method; (c) shows the result of the baseline, which consists of a U-Net for foreground segmentation followed by the original TEASAR algorithm.
Overall, our method accurately captures the vessel structures and aligns well with the ground-truth.
In contrast, the baseline method fails to trace many vessel branches, particularly in the highlighted region within the blue box.
  }
  \label{fig:qualitative_micro-ct}
\end{figure}

\subsection{Ablation experiments in modified TEASAR}\label{sec:ablation_study}
To assess the contribution of each modification to the standard TEASAR algorithm, we conducted a systematic ablation study. Starting from the kimimaro TEASAR, we incrementally add each proposed component in separate experiments. We observe a consistent improvement in performance with every addition in \tableref{tab:ablation_study}, ultimately achieving the lowest false merges and false splits values with our full model configuration.

\begin{table}
    \centering
    \caption{\label{tab:ablation_study}Ablation study illustrating the contribution of individual components of our method and how they incrementally improve performance over a U-Net with standard TEASAR \cite{unet_ronneberger2015,sato2000teasar}.
    We add the following components step by step: support for multiple roots per connected component (multi-root); an additional penalty for tracing along vectors with small magnitudes (vec mag); an additional penalty for tracing in the same direction as the direction vector (vec dir); and adaptive masking to mark processed regions (adapt. mask).
    Results are shown for Vesselpose on the validation set of the Multi-Tree Synthetic dataset.
    }
    \begin{tblr}{width=\linewidth,rows={abovesep=1pt,belowsep=1pt}}
        \midrule
        \SetCell[r=2]{l}{Experiments}
        & \SetCell[r=2]{c}{multi\\root} 
        & \SetCell[r=2]{c}{vec\\mag} 
        & \SetCell[r=2]{c}{vec\\dir} 
        & \SetCell[r=2]{c}{adapt.\\mask}
        & \SetCell[c=3]{c}{Edges} & & &
        \SetCell[c=1]{c}{FM} & \SetCell[c=1]{c}{FS} \\
        \cmidrule[lr]{6-8}
        & & & & & F1$\uparrow$ & Prec$\uparrow$ & Rec$\uparrow$ & Abs.$\downarrow$ & Abs.$\downarrow$\\
        \midrule
        UNet+TEASAR 
        & \xmark & \xmark & \xmark & \xmark
        & 0.46 & 0.63 & 0.36 & 52.4 & 54.1 
        \\
        Ours 
        & \cmark & \xmark & \xmark & \xmark 
        & 0.71 & 0.78 & 0.64 & 38.3 & 38.0 
        \\
         
        & \cmark & \cmark & \xmark & \xmark    
        & 0.70 & 0.78 & 0.64 & 37.3 & 37.6 
        \\
         
        & \cmark & \cmark & \cmark & \xmark    
        & 0.75 & 0.79 & 0.73 & 33.6 & 35.0 
        \\
         
        & \cmark & \cmark  & \cmark & \cmark 
        & 0.80 & 0.81 & 0.79 & 29.7 & 29.3 
        \\
        \midrule
    \end{tblr}
\end{table}

\subsection{Training settings and hyperparameter analysis}\label{sec:training_settings}
\textbf{Training:} We conduct a series of experiments to evaluate the effects of different training settings. Quantitative results for the various datasets are presented in \tableref{tab:training_hyperparameters_mt_syn} and \tableref{tab:training_hyperparameters_Trex-sup}.

\begin{table}
    \centering
    \caption{\label{tab:our_results}Quantitative comparison evaluating the effect of fixed tiles (in a sliding-window fashion) versus random crops from training samples in the multi-tree synthetic dataset.
    }
    \begin{tblr}{width=\linewidth,rows={abovesep=1pt,belowsep=1pt}}
        \midrule
        \SetCell[r=2]{l}{parameters} & 
        \SetCell[c=3]{c}{Edges} & & & 
        \SetCell[c=2]{c}{FM} & & 
        \SetCell[c=2]{c}{FS} & \\
        \cmidrule[lr]{2-4} \cmidrule[lr]{5-6} \cmidrule[lr]{7-8}
        & F1$\uparrow$ & Prec$\uparrow$ & Rec$\uparrow$ & Rel.$\downarrow$ & Abs.$\downarrow$ & Rel.$\downarrow$ & Abs.$\downarrow$\\
        \midrule
        sliding window & 0.79 & 0.80 & 0.78 & 0.008 & 33.36 & 0.008 & 36 \\
        random crops (ours) & 0.80 & 0.81 & 0.79 & 0.007 & 30.33 & 0.007 & 29.3 \\
        \midrule
        \label{tab:training_hyperparameters_mt_syn}
    \end{tblr}
\end{table}

\begin{table}
    \centering
    \caption{\label{tab:our_results}Quantitative comparison evaluating the effect of different training settings—data augmentation, learning rate, and foreground weighting—on single-tree datasets, using Trexplorer-Super metrics \cite{trexplorer_super2025}, consistent with the corresponding baseline studies in \tableref{tab:point_metrics}.
    }
    \begin{tblr}{width=\linewidth,rows={abovesep=1pt,belowsep=1pt},colspec={l c c c c c}}
        \midrule
        \SetCell[r=2]{l}{parameters} & 
        \SetCell[r=2]{c}{Dataset} & 
        \SetCell[c=1]{c}{Point Level} & 
        \SetCell[c=1]{c}{Branch Level} & 
        \SetCell[c=2]{c}{Graph Level} & \\
        \cmidrule[lr]{1-1}\cmidrule[lr]{3-3} \cmidrule[lr]{4-4} \cmidrule[lr]{5-6}
        & & F1$\uparrow$ & F1$\uparrow$ & Betti-0$\downarrow$ & Betti-1$\downarrow$ \\
        \midrule
        no augmentation  & Synthetic  & 88.56 & 80.98 & 0 & 0 \\
        50\% intensity shift (ours)  & Synthetic  & 92.28 & 81.28 & 0 & 0 \\
        \midrule
        learning rate 0.001  & Parse2022 & 36.13 & 23.10 & 1.55 & 0\\
        learning rate 0.0001 (ours)  & Parse2022  & 49.42 & 28.87 & 2.70 & 0 \\
        \midrule
        w/o foreground weight  & Parse2022 & 49.42 & 28.87 & 2.70 & 0 \\
        w/ foreground weight (ours)  & Parse2022 & 57.89 & 36.75 & 1.20 & 0\\
        \midrule
        \label{tab:training_hyperparameters_Trex-sup}
    \end{tblr}
\end{table}

\textbf{TEASAR:} Similarly, we evaluate the TEASAR parameters—the penalty scale (1,000,000) and penalty exponent (16) (cf. \equationref{eq:pv_flow})—with results reported in \tableref{tab:penalty_scale} and \tableref{tab:penalty_exponent}.
Our experiments indicate that varying penalty scale does not lead to substantial changes in TEASAR performance. We therefore retain the value used in the Kimimaro implementation of TEASAR~\cite{Silversmith_Kimimaro_Skeletonize_densely_2021}. This choice is also consistent with the original TEASAR paper~\cite{sato2000teasar}, where this parameter is described as being selected heuristically based on the skeleton segment.
In contrast, penalty exponent has a more pronounced effect on the results. As the exponent increases, the edge-wise F1 score generally improves. However, for very large values, TEASAR begins to merge distinct trees, which is reflected in an increased Betti-1 error. Based on this trade-off, we selected the same value as used in both Kimimaro and the original TEASAR method.
\begin{table}
    \centering
    \caption{Quantitative comparison of our modified TEASAR with varying penalty scale term (cf. \equationref{eq:pv_flow}).
    Results are shown for Vesselpose on the validation data of the Multi-Tree Synthetic dataset.
    We see that results are constant across different penalty scales.
    }
    \begin{tblr}{width=\linewidth,rows={abovesep=1pt,belowsep=1pt}}
        \midrule
        \SetCell[r=2]{l}{penalty scale} & \SetCell[c=3]{c}{Edges} & & & \SetCell[c=2]{c}{FM} & & \SetCell[c=2]{c}{FS} & \\
        \cmidrule[lr]{2-4} \cmidrule[lr]{5-6} \cmidrule[lr]{7-8}
        & F1$\uparrow$ & Prec$\uparrow$ & Rec$\uparrow$ & Rel.$\downarrow$ & Abs.$\downarrow$ & Rel.$\downarrow$ & Abs.$\downarrow$\\
        \midrule
        $5 \times 10^{3}$ & 0.81 & 0.82 & 0.79 & 0.007 & 28 & 0.006 & 27 \\
        $5 \times 10^{4}$ & 0.81 & 0.83 & 0.80 & 0.007 & 29 & 0.007 & 29 \\
        $5 \times 10^{5}$ & 0.80 & 0.82 & 0.79 & 0.007 & 28 & 0.006 & 27 \\
        $5 \times 10^{6}$ & 0.81 & 0.82 & 0.79 & 0.007 & 28 & 0.006 & 27 \\
        $1 \times 10^{6}$ (ours) & 0.81 & 0.82 & 0.79 & 0.007 & 28 & 0.006 & 27 \\
        \midrule
        \label{tab:penalty_scale}
    \end{tblr}
\end{table}

\begin{table}
    \centering
    \caption{Quantitative comparison of our modified TEASAR with varying penalty exponent (cf. \equationref{eq:pv_flow}) on VesselPose validation data from the Multi-Tree Synthetic dataset. Edge-wise F1 score, false merges, and false splits improve with increasing exponent; however, at very high values, TEASAR merges distinct trees, as reflected in the Betti-0 value.}
    \begin{tblr}{width=\linewidth,rows={abovesep=1pt,belowsep=1pt},colspec={l c c c c c c c}}
        \midrule
        \SetCell[r=2]{l}{penalty exp.} & 
        \SetCell[c=1]{c}{Edges} & 
        \SetCell[c=2]{c}{FM} & & 
        \SetCell[c=2]{c}{FS} & &
        \SetCell[c=2]{c}{Betti} & & \\
        \cmidrule[lr]{2-2} \cmidrule[lr]{3-4} \cmidrule[lr]{5-6} \cmidrule[lr]{7-8}
        & F1$\uparrow$ & Rel.$\downarrow$ & Abs.$\downarrow$ & Rel.$\downarrow$ & Abs.$\downarrow$ & Betti-0$\downarrow$ 
        & Betti-1$\downarrow$ \\
        \midrule
        2 & 0.64 & 0.01 & 40 & 0.009 & 39 & 1 & 0 \\
        4 & 0.70 & 0.008 & 33 & 0.007 & 32 & 1 & 0 \\
        8 & 0.77 & 0.008 & 34 & 0.007 & 32 & 1 & 0 \\
        16 (ours) & 0.81 & 0.007 & 28 & 0.007 & 27 & 1 & 0 \\
        32 & 0.81 & 0.005 & 22 & 0.004 & 18 & 4 & 0 \\
        \midrule
        \label{tab:penalty_exponent}
    \end{tblr}
\end{table}

\subsection{Sensitivity to vector prediction quality}\label{sec:vector noise sensitivity}
\begin{figure}[t]
  \centering
  \subfigure[Noise sensitivity plot]{%
    \includegraphics[width=0.81\linewidth]{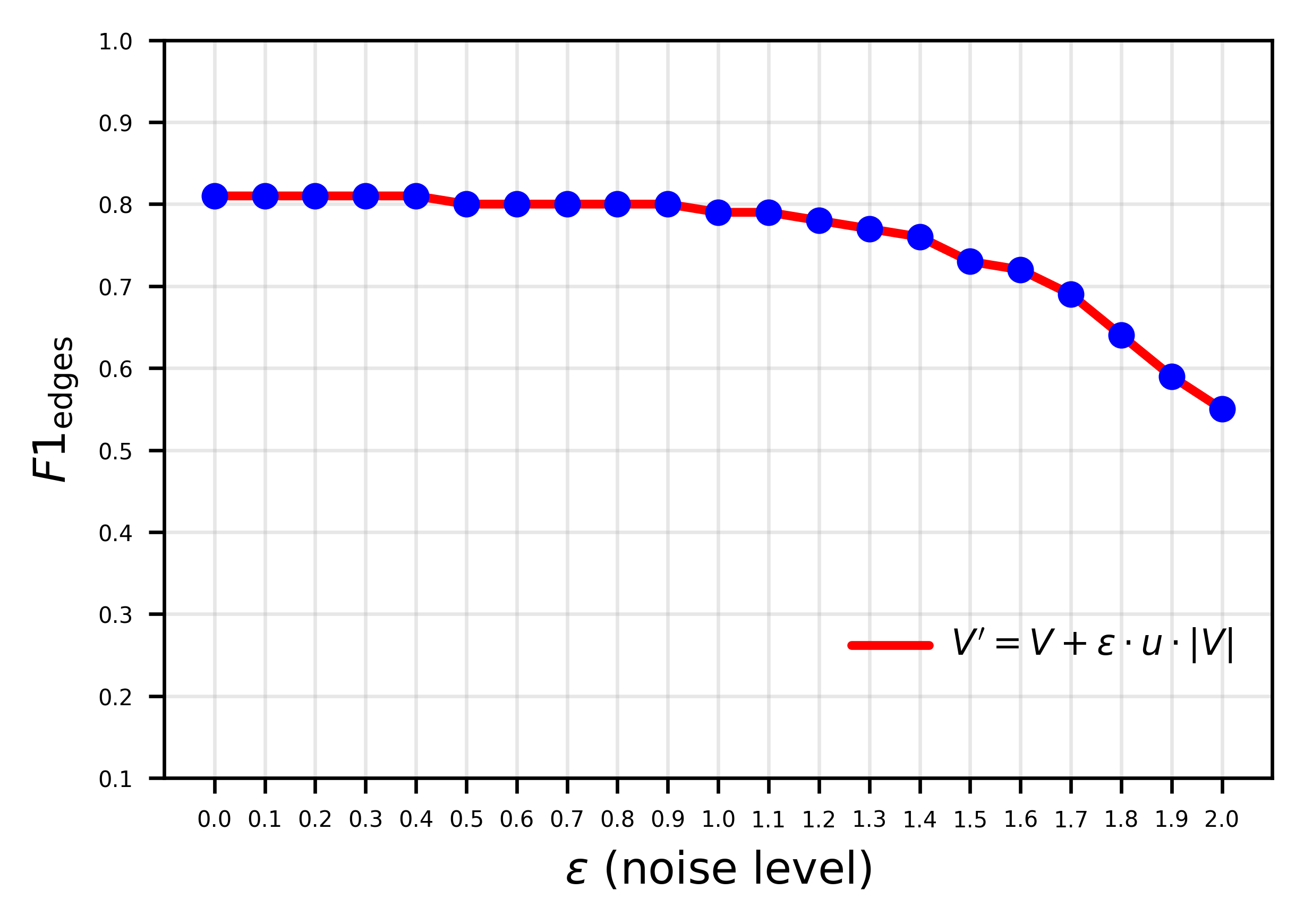}
  }\\
  \subfigure[Original]{%
    \includegraphics[width=0.31\linewidth]{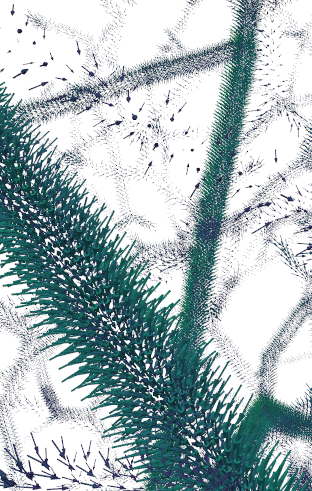}
  }
  \hfill
  \subfigure[$\epsilon=0.5$]{%
    \includegraphics[width=0.31\linewidth]{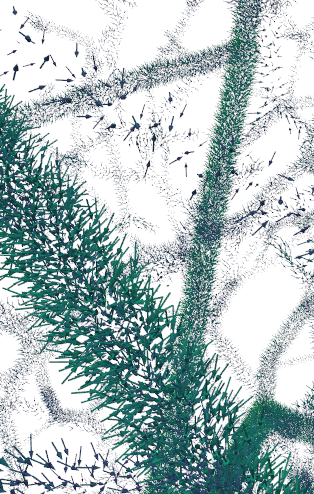}
  }
  \hfill
  \subfigure[$\epsilon=2$]{%
    \includegraphics[width=0.31\linewidth]{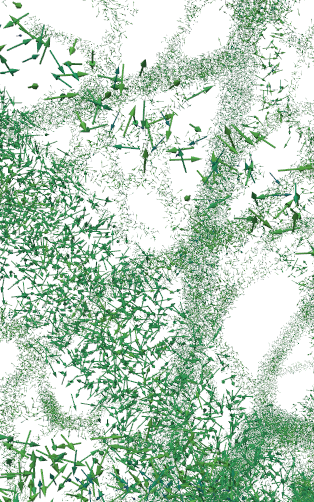}
  }
  \caption{\textbf{Sensitivity to vector noise.}
  (a) The proposed method shows strong robustness to vector noise: small to moderate perturbations (noise level $\varepsilon \leq 1.0$) of the predicted vectors have no noticeable impact on the resulting edge-wise F1 score. (b–d) Visualization of the direction vector field under increasing noise levels  $\varepsilon \in\{0, 0.5, 2\}$. For clarity, all vectors are normalized. Dark blue indicates low vector magnitude, while green indicates high vector magnitude.
  }
  \label{fig:vector_noise sensitivity}
\end{figure}

We evaluate robustness to directional noise by perturbing the predicted vector field with an additive error term such that the error magnitude is proportional to the local predicted vector norm.
Specifically, each original predicted direction vector $v$ is perturbed to $v' = v + \varepsilon \lVert v\rVert\cdot u$, where $\varepsilon \geq 0$ controls the noise level (noise-to-signal ratio) and $u$ is a random unit vector.
We sweep $\varepsilon$ from $0$ to $2.0$ in steps of $0.1$ and quantify performance using the edge-wise $F1$ score. As shown in \figureref{fig:vector_noise sensitivity}(a), our method remains stable over a broad range of perturbation strengths: Edge-wise F1 remains nearly constant for small to moderate noise levels and degrades only gradually as $\varepsilon$ increases. A pronounced drop is observed only at very large noise ($\varepsilon > 1.0$), where the direction field becomes strongly corrupted as seen in \figureref{fig:vector_noise sensitivity}(d) and the reconstruction quality deteriorates more noticeably. Importantly, even for a higher noise level our approach consistently outperforms the baseline \textsc{TEASAR} ($F1_{\mathrm{edge}} = 0.46)$, indicating higher tolerance to directional uncertainty.

\subsection{Sensitivity to test time Gaussian noise}\label{sec:gaussian noise sensitivity}
We further assess robustness to test-time perturbations by adding voxel-wise Gaussian noise to the normalized raw image. Specifically, the noisy input is generated as
\[
I' = \operatorname{clip}(I_{\mathrm{norm}} + \epsilon, 0, 1), \qquad \epsilon \sim \mathcal{N}(0,\sigma^2)
\]
where
\[
I_{\mathrm{norm}} = \frac{I - I_{\min}}{I_{\max} - I_{\min}}
\]
Here, $\sigma$ denotes the standard deviation of the Gaussian noise and controls the noise level. The effect of increasing noise levels during inference is reported in \tableref{tab:gaussian noise sensitivity}. For small values of $\sigma$, the performance remains largely stable, with no substantial degradation. However, at higher noise levels, we observe the emergence of several small disconnected components, which is reflected in the increase in false splits and Betti-0 error.
\begin{table}
    \centering
    \caption{Test time sensitivity study of Vesselpose under varying Gaussian noise.
    Performance remains stable at low $\sigma$, but higher noise levels lead to small disconnected components, increasing false splits and Betti-0 error.
    }
    \begin{tblr}{width=\linewidth,rows={abovesep=1pt,belowsep=1pt},colspec={l c c c c c c c}}
        \midrule
        \SetCell[r=2]{l}{sigma} & 
        \SetCell[c=1]{c}{Edges} & 
        \SetCell[c=2]{c}{FM} & & 
        \SetCell[c=2]{c}{FS} & &
        \SetCell[c=2]{c}{Betti} & & \\
        \cmidrule[lr]{2-2} \cmidrule[lr]{3-4} \cmidrule[lr]{5-6} \cmidrule[lr]{7-8}
        & F1$\uparrow$ & Rel.$\downarrow$ & Abs.$\downarrow$ & Rel.$\downarrow$ & Abs.$\downarrow$ & Betti-0$\downarrow$ 
        & Betti-1$\downarrow$ \\
        \midrule
        0 & 0.81 & 0.007 & 28 & 0.006 & 27 & 1 & 0 \\
        0.03 & 0.81 & 0.007 & 31 & 0.007 & 30 & 1 & 0 \\
        0.06 & 0.81 & 0.008 & 32 & 0.007 & 31 & 1 & 0 \\
        0.09 & 0.80 & 0.008 & 34 & 0.008 & 23 & 1 & 0 \\
        0.12 & 0.80 & 0.01 & 43 & 0.01 & 45 & 2 & 0 \\
        0.15 & 0.78 & 0.009 & 38 & 0.02 & 127 & 94 & 0 \\
        \midrule
        \label{tab:gaussian noise sensitivity}
    \end{tblr}
\end{table}

\subsection{Comparison with nnU-Net}
\label{sec:nnunet}

We evaluate a variant of Vesselpose in which the U-Net backbone is replaced by nnU-Net~\citep{isensee2021nnu}, a self-configuring segmentation framework that automatically determines architecture and training hyperparameters from dataset properties, requiring no manual tuning. We extend the framework to predict voxel-wise direction vectors as additional output channels alongside the foreground mask. The remaining pipeline, including the modified TEASAR algorithm and post-processing, is identical to the main method. Results on the Multi-Tree Synthetic dataset are reported in Table~\ref{tab:nnUNet comaprison}.


\begin{table}
    \centering
    \caption{\label{tab:our_results}Quantitative comparison of Vesselpose with a U-Net versus nnU-Net backbone on the Multi-Tree Synthetic dataset. We observe that nnU-Net performs slightly better on the edge metrics, but shows slightly worse false merge and false split errors than U-Net.}
    \begin{tblr}{width=\linewidth,rows={abovesep=1pt,belowsep=1pt}}
        \midrule
        \SetCell[r=2]{l}{Model} & \SetCell[c=3]{c}{Edges} & & & \SetCell[c=2]{c}{FM} & & \SetCell[c=2]{c}{FS} & \\
        \cmidrule[lr]{2-4} \cmidrule[lr]{5-6} \cmidrule[lr]{7-8}
        & F1$\uparrow$ & Prec$\uparrow$ & Rec$\uparrow$ & Rel.$\downarrow$ & Abs.$\downarrow$ & Rel.$\downarrow$ & Abs.$\downarrow$\\
        \midrule
        Ours(U-Net) & 0.80 & 0.80 & 0.79 & \textbf{0.007} & \textbf{29.7} & \textbf{0.007} & \textbf{28.3} \\
        Ours(nnU-Net) & \textbf{0.81} & \textbf{0.81} & \textbf{0.81} & 0.008 & 33.6 & 0.008 & 32.8\\
        
        \midrule
        \label{tab:nnUNet comaprison}
    \end{tblr}
\end{table}

\end{document}